\documentclass{article}

\usepackage{PRIMEarxiv}

\usepackage{natbib}
\usepackage[utf8]{inputenc} 
\usepackage[T1]{fontenc}    
\usepackage{hyperref}       
\usepackage{url}            
\usepackage{booktabs}       
\usepackage{amsfonts}       
\usepackage{nicefrac}       
\usepackage{microtype}      
\usepackage{lipsum}
\usepackage{fancyhdr}       
\usepackage{graphicx}       
\usepackage[table]{xcolor}  
\usepackage{multirow}
\usepackage{amsmath}
\usepackage{bm}
\usepackage{makecell}
\usepackage{cleveref}
\usepackage{subcaption}
\usepackage{float}
\usepackage{placeins}
\graphicspath{{media/}}     
\newcommand{\namedpar}[1]{\vspace{0.1cm}\textbf{#1}.}

\usepackage[framemethod=default]{mdframed}
\newmdenv[innerlinewidth=0.5pt, roundcorner=4pt,linecolor=blue,innerleftmargin=6pt,
innerrightmargin=6pt,innertopmargin=6pt,innerbottommargin=6pt]{mybox}
\title{Underrepresentation, Label Bias, and Proxies: Towards Data Bias Profiles for the EU AI Act and Beyond} 

\author{
  Marina Ceccon \\
  University of Padova \\
  \texttt{marina.ceccon@phd.unipd.it} \\
   \And
  Giandomenico Cornacchia \\
  IBM Research Europe, Dublin\\
  \texttt{giandomenico.cornacchia1@ibm.com} \\
    \And
  Davide Dalle Pezze \\
  University of Padova \\
  \texttt{davide.dallepezze@unipd.it} \\
\And
  Alessandro Fabris \\
  University of Trieste\\
  MPI-SP, Bochum\\
  \texttt{alessandro.fabris@units.it} \\\\
  \And 
  Gian Antonio Susto \\
  University of Padova \\
  \texttt{gianantonio.susto@unipd.it} \\
  }

\begin{document}
\maketitle

\begin{abstract}
Undesirable biases encoded in the data are key drivers of algorithmic discrimination.
Their importance is widely recognized in the algorithmic fairness literature, as well as legislation and standards on anti-discrimination in AI.
Despite this recognition, data biases remain understudied, hindering the development of computational best practices for their detection and mitigation. 

In this work, we present three common data biases and study their individual and joint effect on algorithmic discrimination across a variety of datasets, models, and fairness measures. 
We find that underrepresentation of vulnerable populations in training sets is less conducive to discrimination than conventionally affirmed, while combinations of proxies and label bias can be far more critical.
Consequently, we develop dedicated mechanisms to detect specific types of bias, and combine them into a preliminary construct we refer to as the \emph{Data Bias Profile (DBP)}. This initial formulation serves as a proof of concept for how different bias signals can be systematically documented.
Through a case study with popular fairness datasets, we demonstrate the effectiveness of the DBP in predicting the risk of discriminatory outcomes and the utility of fairness-enhancing interventions.
Overall, this article bridges algorithmic fairness research and anti-discrimination policy through a data-centric lens. 
\end{abstract}

\keywords{AI Act \and algorithmic fairness \and anti-discrimination \and bias detection \and data bias}

\begin{mybox}
\textbf{Correct reference for this work}:

\vspace{0.2cm}
\begin{small}
\noindent Marina Ceccon, Giandomenico Cornacchia, Davide Dalle Pezze, Alessandro Fabris, Gian Antonio Susto. ``Underrepresentation, label bias, and proxies: Towards Data Bias Profiles for the EU AI act and beyond''. (2025). \textit{Expert Systems with Applications}. 
\noindent \url{https://doi.org/10.1016/j.eswa.2025.128266} 
\end{small}
\end{mybox}

\section{Introduction}
Algorithmic anti-discrimination is a relatively young field, rapidly moving from niche research to market readiness \citep{alverez2024policy}. 
Several years of work carried out by a growing research community have convincingly shown that algorithms developed without attention to fairness put vulnerable groups at a systematic disadvantage \citep{angwin2016machine,obermeyer2019dissecting,glazko2024identifying}. 
Recognizing the critical implications of this research, policymakers and standards organizations have published regulations and norms on the topic \citep{schwartz2022towards,eu2024aiact,iso2021bias}.  
These documents require standardization to apply across many domains where fairness work is critical, including medicine \citep{obermeyer2019dissecting}, finance \citep{gillis2024operationalizing}, employment \citep{fabris2024fairness}, and education \citep{baker2022algorithmic}.

Evaluations of data bias are key computational tools for anti-discrimination work.
Since biases in the data are fundamental drivers of algorithmic discrimination \citep{vetro2021data,brzezinski2024properties}, 
bias management is mentioned in every recent standard and regulation on algorithmic anti-discrimination \citep{schwartz2022towards,eu2024aiact,iso2021bias}. 
Policy formulations on this topic are rather vague, favouring flexibility on one hand, but leaving the contours of law-abiding bias management undefined for practitioners, contributing to legal risk and uncertainty. 
For example, recent regulation requires providers of AI systems in high-risk domains to signal sufficient efforts of bias detection and mitigation. How this should be done in practice, however, is left completely undefined \citep{deck2024implications}.

Defining precise criteria for bias management requires answering several important questions, left mostly unaddressed in the fairness literature. First, different data biases are associated with algorithmic discrimination.  Which combinations of biases are more critical for fairness? Second, data biases are typically described qualitatively but vastly lack a quantitative characterization. Is it possible to monitor distinct data biases unambiguously? Third, while a plethora of fairness interventions exist in the literature, a set of guidelines to prioritize them is prominently lacking. In sum, how can practitioners and researchers make decisions about bias mitigation in a principled manner?

\namedpar{Contributions} In this work, we tackle the above questions, providing several contributions.
\begin{itemize}
    \item We study three types of data bias widely recognized for their negative influence on algorithmic fairness, namely underrepresentation, label bias, and proxies. Through extensive experiments on diverse datasets, algorithms, and fairness measures we analyze their joint influence on algorithmic discrimination. Our experiments show that underrepresentation in training data is overemphasized in the literature while label bias is more critical. This novel result challenges conventional wisdom held in the algorithmic fairness community.
    \item We propose a principled mechanism for bias detection that is widely applicable in practice. More in detail, we develop a suite of measures to detect specific data biases without auxiliary information from external sources. We integrate these measures into a preliminary construct, the \emph{Data Bias Profile (DBP)}, which provides a quantitative foundation for identifying and communicating data biases, as well as assessing the risk of algorithmic discrimination. This framework serves as a proof of concept, offering a concrete starting point that the research community can build upon to develop a more robust and systematic approach to bias-aware data documentation.   
    \item We discuss the far-reaching implications of our work for researchers and practitioners. We recommend that researchers use DBPs to select datasets with complementary properties for their experiments, overcoming the present limitations of fairness benchmarks. Finally, we make recommendations for practitioners on the curation and utilization of anti-discriminatory datasets.
\end{itemize}

\namedpar{Structure} Section \ref{sec:related} presents related work. Section \ref{sec:3_databias} introduces data bias and experimental protocols to analyze it. Section \ref{sec:3_experiments} studies the effect of data bias. Section \ref{sec:bias_detection} describes bias detection, introducing and demonstrating  DBPs. Sections \ref{sec:discussion} and \ref{sec:concl} discuss our results in the broader context of algorithmic fairness work concluding with recommendations for researchers and practitioners.

\section{Related Work}\label{sec:related}
Algorithmic fairness research keeps contributing new approaches to measure the risk of discrimination \citep{cornacchia2023auditing,chen2024measuring,fabris2023pairwise,cooper2024arbitrariness} and to mitigate it \citep{yin2023longterm,cruz2024unprocessing,zhang2024learning,fajri2024falcur}. Moving fairness toward market readiness requires research on operationalizing algorithmic anti-discrimination policy (\cref{sec:related_policy}), on its close connection with data bias (\cref{sec:related_databias}), and on principled documentation practices to support anti-discrimination (\cref{sec:related_documentation}).   

\subsection{AI policy on anti-discrimination}\label{sec:related_policy}
Influential legislation and standards on anti-discrimination in AI, such as the EU AI Act \citep{eu2024aiact} and the NIST report on bias in AI \citep{schwartz2022towards} require multi-disciplinary research to translate policy into computational best practices. 
\citet{drukker2023toward}, for example, complement NIST guidelines with a list of domain-specific biases that arise in the medical domain.
\citet{borgesius2024non} focus on the AI Act, summarizing the main requirements for mandatory evaluation of data biases and their documentation.
\citet{deck2024implications} compile a list of practical challenges for compliance with anti-discrimination requirements outlined in the AI Act. By highlighting the need for model owners to examine ``possible biases that are likely to [...] lead to discrimination prohibited under Union law'', they stress the pressing questions of which biases lead to discriminatory models and what kind of evidence is required to signal sufficient efforts for bias detection and correction.  Our work aims to provide flexible computational tools to answer this question across different domains.

\subsection{Linking fairness with data properties}\label{sec:related_databias}

A growing line of work centers on quantifying data biases and their influence on models.
\citet{guerdan2023groundless} describe five sources of bias affecting target variables and develop a causal framework to disentangle them. \citet{baumann2023bias} present several data biases and provide initial insights into their mitigation. \citet{brzezinski2024properties} study the variability of fairness measures with respect to the underrepresentation of protected groups and the imbalance between positives and negatives. They postulate certain properties (e.g. ``fairness should not vary with underrepresentation'') and highlight measures that realize said properties as more ``reliable''. \citet{fragkathoulas2024explaining} survey the intersection of fairness and explainability, including explanations that can describe sources of unfairness.

\citet{vetro2021data} set out to predict the risk of discrimination against vulnerable groups from their underrepresentation in the data. Their work is highly influential for ours; it represents a first attempt at developing mechanisms to detect (a single type of) data bias and connect it with model fairness, opening the way to follow-up studies \citep{mecati2022detecting,mecati2023measuring}. Our manuscript continues this line of work, with important differences in methodology and conclusions. First, we consider three types of data bias, adding label bias and proxies to underrepresentation and studying their joint influence on fairness. Second, we assess model fairness on unbiased test sets. For example, to measure the effect of strong under-representation, we remove from the training set a large percentage of items from the disadvantaged group (even 100\%), but we retain them in the test set for a reliable evaluation. Third, we conclude that underrepresentation in training sets is overemphasized in the algorithmic fairness literature and that other data biases can be more critical.

\subsection{Quantitative data documentation}\label{sec:related_documentation}
Data documentation is increasingly recognized as a central component of trustworthy AI \citep{gebru2021datasheets,holland2020dataset,pushkarna2022data,fabris2022algorithmic,konigstorfer2022ai,rondina2023completeness,golpayegani2024ai,sambasivan2021everyone}. 
With few exceptions, prominent data documentation frameworks are qualitative.
Among quantitative frameworks, \citet{holland2020dataset} emphasize the analysis of correlations between variables to spot anomalous trends. 
\citet{dominguez2024metrics} develop metrics to quantify representational and stereotypical biases, demonstrating them on a facial expression recognition dataset. 
In this work, we propose a principled suite of measures to quantify and document biases associated with algorithmic discrimination. We then outline how these can be composed into a preliminary construct, the Data Bias Profile (DBP), to support structured documentation and analysis. Our approach is tailored to one specific aspect of datasets and differs from existing methods, both quantitative and qualitative.

\section{Data Bias}\label{sec:3_databias}
Data is a fundamental driver of algorithmic discrimination \citep{mehrabi2021survey,suresh2021framework,schwartz2022towards}. Data biases are defined as data properties that, if unaddressed, lead to AI systems that perform better or worse for different groups \citep{iso2021bias}. In this section, we describe three types of data bias widely recognized for their impact on algorithmic fairness along with corresponding bias injection mechanisms.

\subsection{Underrepresentation}\label{sec:underrepr}
The term \emph{representativeness} typically refers to the ability of a dataset to support the development of an accurate model for a target population. 
 Underrepresentation of disadvantaged groups in data is described at length in popular media \citep{perez2019invisible,cobham2020uncounted} and seminal fairness articles \citep{shankar2017no,buolamwini2018gender,mehrabi2021survey} as a key driver of algorithmic discrimination. When groups from the target population are underrepresented in training data, it is argued, AI models will fail to generalize and underperform for those groups \citep{suresh2021framework}. Indeed, the (under)representation of protected groups in training sets is studied as a predictor of model unfairness \citep{vetro2021data,brzezinski2024properties}. Influential legislation and standards recognize representation as a central component of algorithmic anti-discrimination  \citep{eu2024aiact,schwartz2022towards} and mandate efforts to document and curb it.

\subsection{Label bias}\label{sec:label_bias}

Labels (or target variables) are key to AI. They give machine learning a ``ground truth'' that models learn to replicate. Since data is a social mirror, labels reflect undesirable disparities in society \citep{barocas2023fairness}. Indeed, measurement methods can be biased across protected groups \citep{vardasbi2024impact}. Policing and arrest tend to target poorer neighborhoods, therefore biasing crime data against black US citizens \citep{bao2021compaslicated}. Medical data suffers from underdiagnosis due to substandard medical care \citep{gianfrancesco2018potential} and barriers to access for vulnerable populations \citep{obermeyer2019dissecting}. Semi-automated labels are especially likely to compound and reinforce spurious biases in training datasets \citep{jigsaw2018unintended}. Several methods have been proposed in the literature to counteract unfair label biases under simplifying assumptions \citep{kamiran2011preprocessing,feldman2015certifying,wang2021fair,liu2024aim}. Overall, measurement bias is inevitable \citep{jacobs2021measurement}; it is especially problematic for anti-discrimination when it tilts target labels against a vulnerable group \citep{mehrabi2021survey,suresh2021framework}. Models trained to predict these labels will encode the underlying biases and harm disadvantaged groups \citep{obermeyer2019dissecting,bao2021compaslicated}.

\subsection{Proxies}
Proxies are features that correlate with protected attributes. Protected attributes such as gender can be revealed by individual features, such as names in a resume \citep{santamaria2018comparison}, or by combinations of features, such as the browsing history of a person \citep{hu2007demographic}. Pursuing fairness by simply removing protected attributes from input features, an approach termed \emph{fairness through unawareness}, is ineffective precisely for this reason: a redundant encoding of latent protected variables is present in other features \citep{hardt2016equality,pedreschi2008discrimination,barocas2016big}. Proxy removal, for example through feature selection or projection, is a popular approach to improve algorithmic fairness \citep{madras2018learning,edwards2016censoring,alves2021reducing,hireview2022explainability,blindstairs2024unbiased}. Conversely, input features that are strongly correlated with protected attributes are considered a driver of unfairness in data-driven models \citep{schwartz2022towards}. Policymakers may therefore expect practitioners to actively identify and eliminate strong proxy features from models powering automated decisions \citep{bogen2024navigating}.

\section{Effect of Data Bias}\label{sec:3_experiments}
In this section, we investigate the effect of data biases.
We inject data bias into the training and validation datasets of classification models and assess their combined influence on algorithmic discrimination, evaluating fairness metrics on unbiased test sets.
\subsection{Overall setup}
\begin{table}[!htp]
\centering
\caption{\textbf{Dataset basics}. We report dataset name, sensitive attribute information such as (dis)advantaged groups and their prevalence, and target variables information such as positive classes and their prevalence among members of the advantaged and disadvantaged groups.} 
\label{tab:DatasetsStatistics}
\tiny
\renewcommand{\arraystretch}{1.5}
\setlength{\tabcolsep}{8pt}
\begin{tabular}{p{1.4cm}p{0.8cm}p{0.8cm}p{0.8cm}p{0.9cm}p{1.0cm}p{1.0cm}p{1.0cm}p{1.0cm}} \toprule
\textbf{Dataset} &$\bm{s}$ &$\bm{a}$ &$\bm{d}$ & $\bm{y}$ &$\bm{y=\oplus}$ & \scalebox{.95}{$\bm{\Pr}_{\sigma}(s=a)$}& \scalebox{.95}{\makecell{$\bm{\Pr}_{\sigma}(y=\oplus$ \\ $\mid s=a)$}} & \scalebox{.95}{\makecell{$\bm{\Pr}_{\sigma}(y=\oplus$ \\ $\mid s=d)$}} \\
\hline
\rowcolor{lightgray} &Gender &Male &Female  & & &0.68&0.3125 &0.1136\\
\rowcolor{lightgray}\multirow{-2}{*}{Adult} &Marital status &Married &Not married/ Divorced & \multirow{-2}{*}{income} & \multirow{-2}{*}{$>$ 50K}& 0.48 &0.4451 &0.0668 \\
Compas&Ethnicity &Caucasian &African-American & recidivism & no reoffense&0.60 &0.6091 & 0.4769 \\
\rowcolor{lightgray}
Crime &Ethnicity &Caucasian &Other & violent crime rate & low  &0.58 &0.7335 &0.1805 \\
Folktables &Ethnicity &Caucasian &Other & employment & employed &0.89 &0.5688 &0.5048 \\
\rowcolor{lightgray}
German &Age &$>$ 25 y &$<=$ 25 y & credit risk& good &0.81&0.7284 &0.5789 \\
NIH &Gender &Male &Female & chest pathologies & presence of pathology &0.54&0.4112 &0.3981 \\
\rowcolor{lightgray}
Fitzpatrick17k &Skin type & Light & Dark & skin conditions & presence of condition &0.86&0.2826 &0.1897 \\
\bottomrule
\end{tabular}
\end{table}

\definecolor{lightgray}{gray}{0.95}
\begin{table}[H]
\centering
\caption{\textbf{Notation}. Main notational convention adopted in this work.}\label{tab:notation}
\footnotesize
\rowcolors{1}{white}{lightgray}
\begin{tabular}{p{2.4cm}p{6cm}}
\hiderowcolors
\toprule
\textbf{symbol} & \textbf{meaning}  \\
\hline
\showrowcolors
$s \in \mathcal{S}$ &  protected attribute \\
$s=d$ &  historically disadvantaged group \\
$s=a$ &  historically advantaged group \\
$y \in \mathcal{Y}$ &  target variables \\
$\mathcal{Y}=\{ \oplus, \ominus \}$ &  positive and negative target values \\
$x \in \mathcal{X}$ &  non-protected attributes \\
$\hat{y}=g(x)$ & estimation of $y$ through classifier $g(\cdot)$ \\
$\sigma=\{(x_i, y_i, s_i)\}$ & a training set \\
$\Pr_{\sigma}(s=d)$ & prevalence of $d$ in set $\sigma$ \\
$\sigma_d =\{ i \in \sigma | s_i=d \}$ & subset of $\sigma$ with all data points from group $d$ \\
$\sigma', y'$ & training set and target labels after bias injection \\
$r \in (0,1)$ & percentage of disadvantaged instances retained for training: $\Pr_{\sigma'}(s=d)=r \cdot \Pr_{\sigma}(s=d)$\\
$u=r-1$ & underrepresentation factor \\
$f \in (0,1)$ & flip factor or label bias: $f = \Pr(y_i'=\ominus|y_i=\oplus, s_i=d)$ \\
\bottomrule
\end{tabular}
\end{table}

\subsection{Data bias injection} \label{sec:bias_inject_method}
\namedpar{Notation} 
Table \ref{tab:notation} summarizes the notational conventions.
We let $s \in \mathcal{S}$ denote a sensitive attribute,\footnote{We use the nomenclature \emph{sensitive} and \emph{protected} attribute interchangeably.} with value $s=a$ ($s=d$) denoting the historically (dis)advantaged group. We let $y$ indicate the target variable with values in  $\mathcal{Y}=\{ \oplus, \ominus \}$ and we let $x \in \mathcal{X}$ denote the non-protected features used for classification. Target variables are estimated through a classifier $\hat{y}=g(x)$. We let $\sigma=\{(x_i, y_i, s_i)\}$ denote a sample and $\Pr_{\sigma}(s=d)$ indicate the prevalence of items with $s_i=d$ in that sample. To inject biases in training sets, we subsample the disadvantaged group and flip its labels. We use $\sigma'$ to denote a training set derived from $\sigma$ via bias injection.  


\namedpar{Underrepresentation} We let $r \in (0,1)$ denote the percentage of instances from the disadvantaged group retained for training, so that 
\begin{align}
    \Pr_{\sigma'}(s=d) &=r \cdot \Pr_{\sigma}(s=d) \nonumber \\
    u &= r-1.
\end{align}
We call $u=1-r$ the \emph{underrepresentation factor} for the disadvantaged group. We vary $u$ across its full range; extreme values $u=1$ and $u=0$ denote complete underrepresentation and no underrepresentation, respectively. 

\namedpar{Label bias} For label bias, we selectively flip labels. We let $f \in (0,1)$ indicate the proportion of positive instances ($y=\oplus$) from the vulnerable group whose label is flipped to negative ($y=\ominus$), i.e. 
\begin{align}
    f&=\Pr(y_i'=\ominus|y_i=\oplus, s_i=d). \label{eq:flip}
\end{align}
We let the \emph{flip factor} $f$ vary between $f=0$ and $f=1$; the former corresponds to no bias injection, the latter to maximum bias where all the positive items from the disadvantaged group in the training set are flipped to a negative target label. 

\namedpar{Proxies} We quantify the strength of proxies as their joint ability to predict sensitive attributes. We train a classifier $\hat{s}=h(x)$ to estimate the protected attribute $s$ and we compute its AUC to measure the strength of proxies. 
\begin{align}
    \hat{s}&=h(x) \nonumber \\
    s\text{AUC} &= \text{AUC}(h). \label{eq:sauc_def}
\end{align}
We term $s\text{AUC}$ the \emph{proxy factor} and propose two mechanisms to vary it. 
An additive protocol adds to the non-sensitive variables $\mathcal{X}$ a new feature correlated with sensitive variables
\begin{align}
    x_{\text{new}} &= s + v, \quad v \sim \mathcal{N}(0, \text{std}^2) \nonumber \\
    \mathcal{X}' &= \mathcal{X} \times \mathcal{X}_{\text{new}},
\end{align}
where $v$ is a normal random variable with zero mean and $\text{std}^2$ variance; we increase the strength of proxies by reducing $\text{std}$. In addition, we consider a subtractive protocol, iteratively removing from the non-sensitive variables $\mathcal{X}$ the strongest predictors of the sensitive attribute
\begin{align}
    x_{\text{drop}} &= \max_x \text{sim} (x, s) \nonumber \\
    \mathcal{X}' &= \mathcal{X} \setminus \mathcal{X}_{\text{drop}}, \label{eq:prot_prox_drop}
\end{align}
where $\text{sim}(\cdot, \cdot)$ denotes a similarity function (e.g. correlation) and $\mathcal{X}  \setminus  \mathcal{X}_{\text{drop}}$ is the feature space obtained removing $x_{\text{drop}}$ from the original feature set.\footnote{Equation \eqref{eq:prot_prox_drop} is a single iteration of the subtractive protocol.}

\namedpar{Datasets}
We consider five tabular and two medical imaging datasets, described in Table \ref{tab:DatasetsStatistics}. These datasets are popular in the fairness literature and span several domains where fairness work is critical.
Datasets contain information on protected attributes including gender, age, ethnicity, and marital status. Table \ref{tab:DatasetsStatistics} additionally reports the target variable of each dataset, the prevalence of the disadvantaged and the advantaged groups, as well as the prevalence of positive items in each group. The advantaged group has a higher rate of positive samples compared to the disadvantaged group. Notice that the positive class indicates more desirable outcomes for the assessed individuals, insofar as it is associated with critical resource allocation (loans, medical attention) or lower penalties (incarceration, strict policing). This makes high true positive rates unambiguously important to counter undesirable patterns harming disadvantaged groups, such as underallocation and overcriminalization.
More details on each dataset are reported in \ref{app:data}.

\namedpar{Models}
We train deep learning models for medical imaging datasets and traditional machine learning models for tabular data. Models optimize accuracy-oriented loss functions without any fairness-enhancing component. For each of the tabular datasets, we train a random forest (RF), a support vector classifier (SVC), and a linear regression (LR). Following the literature, we train a Densenet121 on NIH and a vgg16 model on Fitzpatrick17k \citep{seyyedkalantari2020chexclusion, groh2021evaluating}.

\textbf{Splits \& repetitions}. We process tabular datasets with an 80-10-10 train-validation-test split. For NIH, we follow the literature with an 80-10-10 train-validation-test split \citep{seyyedkalantari2020chexclusion}. For Fitzpatrick17k, we use a 70-15-15 split to ensure sufficient representation of the disadvantaged group in the test set, favoring more stable fairness measurements. After splitting the data, we inject biases in the training and validation set. We keep the test set \emph{unbiased} for a reliable evaluation.\footnote{We use the term \emph{unbiased} in a narrow sense, to denote simple random samples of the original dataset, as opposed to subsets where we deliberately inject different types of data bias.} 
For each experiment, we perform 10 training repetitions (with different initial seeds), reporting the mean and standard deviation for metrics of interest. 

\namedpar{Performance Metrics} To evaluate the classification performance of each model across (often imbalanced) datasets, we consider the balanced accuracy on the test set, i.e. the average between the true positive rate and the true negative rate:
\begin{equation}
\text{BA}_{\sigma} = \frac{{\Pr}_{\sigma}(\hat{y}=\oplus \mid y=\oplus) + {\Pr}_{\sigma}(\hat{y}=\ominus \mid y=\ominus)}{2}.
\end{equation}

\namedpar{Fairness Metrics}
To assess the model fairness, we consider three complementary metrics: demographic parity, equal opportunity, and predictive quality parity. Demographic parity (DP), also called independence \citep{barocas2023fairness}, and instantiated as a mean difference \citep{zliobaite2017measuring}, is defined as the difference between the acceptance rates computed on different groups 
\begin{equation}
    \text{DP}_{\sigma} = {\Pr}_{\sigma}(\hat{y}=\oplus \mid s=a) - {\Pr}_{\sigma}(\hat{y}=\oplus \mid s=d) \label{eq:dp}
\end{equation}
and it is independent of the ground truth labels. It is especially salient in contexts where reliable ground truth information is hard to obtain and a positive outcome is desirable, including employment, credit, and criminal justice \citep{du2020fairness,gajane2018formalizing}.

Contrary to DP, the equal opportunity (EO) metric is based on the target variable $y$ \citep{hardt2016equality}; it is defined as the difference in the true positive rates:
\begin{equation}
    \text{EO}_{\sigma} = {\Pr}_{\sigma}(\hat{y}=\oplus \mid s=a, y = \oplus) - {\Pr}_{\sigma}(\hat{y}=\oplus \mid s=d, y = \oplus). \label{eq:eo}
\end{equation}
EO is especially important in contexts, such as healthcare, where a ground truth of reasonable accuracy is available and false negatives (missed diagnosis) are especially harmful.

A third anti-discrimination criterion, focused on both types of misclassification, is represented by prediction quality parity (PQP) \citep{du2020fairness}. We define it as the difference in balanced accuracy between sensitive groups:
\begin{equation}
    \text{PQP}_{\sigma} = \text{BA}_{\sigma_a} - \text{BA}_{\sigma_d}.
\end{equation}
In the experiments below, we measure algorithmic fairness according to these metrics as we inject controlled biases the training sets. 
We present results for logistic regression (on tabular datasets) and equal opportunity, which are representative of broader trends across all models and metrics.
The results for the remaining models and metrics can be found in  \ref{app:exp_effect_bias}; unless explicitly stated, they are equivalent to those illustrated below.
\subsection{Underrepresentation}\label{sec:underrepresentation}

\namedpar{Setup} To study the effect of underrepresentation, we train models in four different settings, varying the underrepresentation
factor by undersampling the disadvantaged group. 
We consider the unbiased case ($u=0$), the fully biased case ($u=1$), where the disadvantaged group is completely absent from the training set, and two intermediate settings ($u=0.2$, $u=0.8$).

\begin{table}[!htp]\centering
\caption{\textbf{Model fairness is mostly unaffected by underrepresentation in the training set}. Equal Opportunity ($\text{EO}$), varying the underrepresentation of the minority group in the training set from $u=0$ (no bias) to $u=1$ (maximum bias). Mean and standard deviation over 10 repetitions. Symbols (*) and (**) denote statistically significant differences with respect to $u=0$ at $p=0.05$ and $p=0.01$, respectively, measured with an unpaired $t$-test.}\label{tab:deltaTPrUnbal }
\tiny
\renewcommand{\arraystretch}{1.5}
\begin{tabular}{llllllll}\toprule
& & & \multicolumn{4}{c}{$\text{EO}$} \\ \cmidrule(lr){4-7}
\textbf{Dataset} & \textbf{sensitive} & \textbf{model} & \multicolumn{1}{c}{$u=0$} & \multicolumn{1}{c}{$u=0.2$} & \multicolumn{1}{c}{$u=0.8$} & \multicolumn{1}{c}{$u=1$} \\ 
& & & \multicolumn{1}{c}{\tiny{(no bias)}} & \multicolumn{1}{c}{} & \multicolumn{1}{c}{} & \multicolumn{1}{c}{\tiny{(max bias)}} \\ \midrule
 \rowcolor{lightgray} &gender &
LR &0.08 $\pm$ 0.02\phantom{**} &0.09 $\pm$ 0.02\phantom{**} &0.09 $\pm$ 0.02\phantom{**} &0.21 $\pm$ 0.07** \\
 \rowcolor{lightgray}\multirow{-2}{*}{Adult} &marital-status &
LR &0.36 $\pm$ 0.03\phantom{**} &0.36 $\pm$ 0.03\phantom{**} &0.37 $\pm$ 0.03\phantom{**} &0.29 $\pm$ 0.04** \\
 Compas &race &
LR &0.17 $\pm$ 0.03\phantom{**} &0.17 $\pm$ 0.03\phantom{**} &0.16 $\pm$ 0.02\phantom{**} &0.16 $\pm$ 0.02\phantom{**} \\
 \rowcolor{lightgray}Crime &race &
LR &0.33 $\pm$ 0.11\phantom{**} &0.36 $\pm$ 0.12\phantom{**} &0.30 $\pm$ 0.14\phantom{**} &0.28 $\pm$ 0.13\phantom{**} \\
Folktables &race &LR &0.05 $\pm$ 0.04\phantom{**} &0.05 $\pm$ 0.04\phantom{**} &0.05 $\pm$ 0.04\phantom{**} &0.05 $\pm$ 0.04\phantom{**} \\
\rowcolor{lightgray} German &age &
LR &0.07 $\pm$ 0.13\phantom{**} &0.06 $\pm$ 0.11\phantom{**} &0.07 $\pm$ 0.09\phantom{**} &0.06 $\pm$ 0.07\phantom{**} \\
NIH &gender &DenseNet &0.01 $\pm$ 0.02\phantom{**} &0.04 $\pm$ 0.01** &0.03 $\pm$ 0.02\phantom{**} &0.06 $\pm$ 0.02** \\
\rowcolor{lightgray}Fitzpatrick17k &skin type &vgg16 &0.09 $\pm$ 0.02\phantom{**} &0.11 $\pm$ 0.02\phantom{**} &0.11 $\pm$ 0.02\phantom{**} &0.11 $\pm$ 0.02\phantom{**} \\
\bottomrule
\label{tab:TPR_gap_underrepresentation}
\end{tabular}
\end{table}

\begin{figure}
    \centering
    \includegraphics[width=\linewidth]{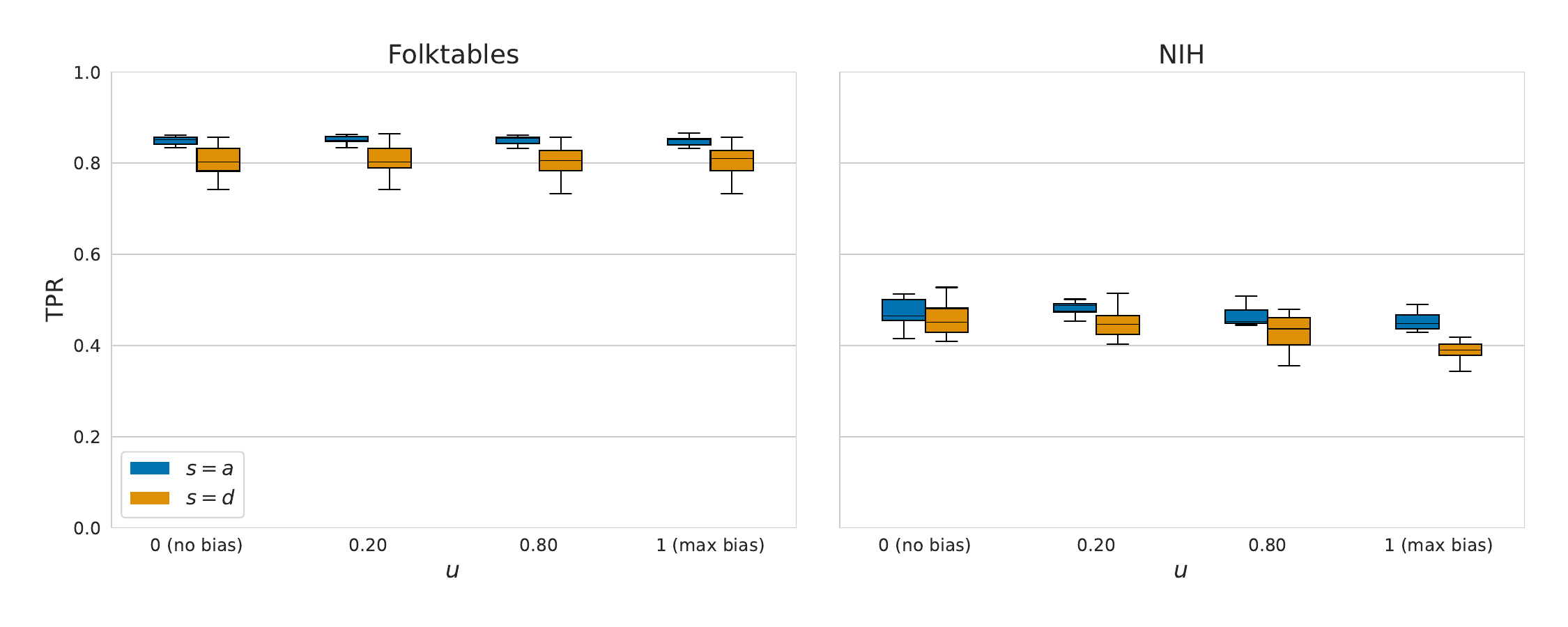}
    \caption{\textbf{Large underrepresentation induces minor variations in the True positive rates (TPR) of both groups}. Boxplots represent the TPR of the advantaged ($s=a$) and disadvantaged group ($s=d$), as u varies.}
    \label{fig:TPR_gap_underrepresentation}
\end{figure}

\namedpar{Results}
Remarkably, Table \ref{tab:TPR_gap_underrepresentation} shows that the underrepresentation of the minority group does not have a strong impact on fairness: EO is approximately constant as $u$ varies in all datasets. 

The only datasets for which the increase in disparity is statistically significant are NIH and Adult (gender). For Crime and Adult (marital status), the gap even decreases slightly when the disadvantaged group is removed from the training set.

This trend is surprising and contradicts popular narratives about the effect of underrepresentation on algorithmic fairness. To further analyze these results, we split EO into its groupwise TPR components (Equation \ref{eq:eo}). Figure \ref{fig:TPR_gap_underrepresentation} reports boxplots of the TPR for the advantaged and disadvantaged groups in NIH and Folktables, which are representative of the remaining datasets.
Figure \ref{fig:TPR_gap_underrepresentation} shows that the TPR remains approximately stable as underrepresentation $u$ varies maximally. Specifically, the TPR for both the advantaged and disadvantaged groups in Folktables are perfectly stable, while they slightly decrease for NIH. The decrease is more marked for the disadvantaged group, leading to a small increase in EO. Underrepresentation is more impactful for NIH since nearly half of the original training set consists of points from the disadvantaged group (Table \ref{tab:DatasetsStatistics}).

\namedpar{Interpretation} This notable result contradicts the position commonly held in algorithmic fairness that increasing the representation of disadvantaged groups in training sets is critical for equitable outcomes. We defer a broader interpretation of this result to Section \ref{sec:discussion}, where we discuss our findings in the broader context of algorithmic fairness research and practice. For now, we highlight this as an indication that underrepresentation in training sets is overemphasized and that other biases may be stronger drivers of model unfairness.

\subsection{Label bias}\label{sec:label_bias_exp}
\namedpar{Setup} In this section, we train models on data affected by different degrees of label bias. Specifically, we take a portion ($f$) of positive samples from the disadvantaged group in the training set and flip their labels to negative. We let the flip factor (Equation \ref{eq:flip}) take values $f=0$ (no bias), $f=0.2$ (moderate bias), $f=0.8$ (strong bias), and $f=1$ (maximum bias).

Additionally, we study the interplay between label bias and underrepresentation by analyzing a scenario in which the prevalence of the disadvantaged group is decreased and part of its positives are flipped. More in detail, at the end of this section, we assess the joint effect of a weak label bias ($f \in \{0, 0.2\}$) and widely-ranging underrepresentation ($u \in \{0,1 \}$).

We report the mean and standard deviation of $\text{EO}$ across ten repetitions. As in the previous section, we focus on LR, while results for other models are available in \ref{app:exp_effect_bias} along with additional fairness measures.

\begin{table}[!htp]\centering
\caption{\textbf{Model fairness is strongly affected by label bias}. Equal Opportunity ($\text{EO}$), as the percentage of flipped positives in the disadvantaged group varies from $f=0$ (no bias) to $f=1$ (maximum bias). Mean and standard deviation over 10 repetitions. Symbols (*) and (**) denote statistically significant differences with respect to $f=0$ at $p=0.05$ and $p=0.01$, respectively, measured with an unpaired $t$-test.}
\label{tab:deltaTPrLabelBias}
\tiny
\renewcommand{\arraystretch}{1.5} 
\begin{tabular}{llllllll}\toprule
& & & \multicolumn{4}{c}{$\text{EO}$} \\ \cmidrule(lr){4-7}
\textbf{Dataset} & \textbf{sensitive} & \textbf{model} & \multicolumn{1}{c}{$f=0$} & \multicolumn{1}{c}{$f=0.2$} & \multicolumn{1}{c}{$f=0.8$} & \multicolumn{1}{c}{$f=1$} \\ 
& & & \multicolumn{1}{c}{\tiny{(no bias)}} & \multicolumn{1}{c}{} & \multicolumn{1}{c}{} & \multicolumn{1}{c}{\tiny{(max bias)}} \\ \midrule
\rowcolor{lightgray} &gender &
LR &0.08 $\pm$ 0.02\phantom{**} &0.21 $\pm$ 0.02** &0.52 $\pm$ 0.03** &0.55 $\pm$ 0.02** \\
\rowcolor{lightgray} \multirow{-2}{*}{Adult} &marital-status &
LR &0.36 $\pm$ 0.03\phantom{**} &0.43 $\pm$ 0.04** &0.62 $\pm$ 0.02** &0.63 $\pm$ 0.02** \\
 Compas  &race &
LR &0.17 $\pm$ 0.03\phantom{**} &0.21 $\pm$ 0.05\phantom{**} &0.21 $\pm$ 0.05\phantom{**} &0.15 $\pm$ 0.05\phantom{**} \\
\rowcolor{lightgray} Crime &race &
LR &0.33 $\pm$ 0.11\phantom{**} &0.39 $\pm$ 0.11\phantom{**} &0.62 $\pm$ 0.17** &0.67 $\pm$ 0.15**\\
 Folktables &race &
LR &0.05 $\pm$ 0.04\phantom{**} &0.05 $\pm$ 0.04\phantom{**} &0.08 $\pm$ 0.04\phantom{**} &0.09 $\pm$ 0.04\phantom{**} \\
\rowcolor{lightgray} German &age &
LR &0.07 $\pm$ 0.13\phantom{**} &0.10 $\pm$ 0.14\phantom{**} &0.22 $\pm$ 0.16\phantom{**} &0.25 $\pm$ 0.10**\\
 NIH &gender &DenseNet &0.01 $\pm$ 0.02\phantom{**} &0.09 $\pm$ 0.02** &0.40 $\pm$ 0.03** &0.48 $\pm$ 0.01** \\
\rowcolor{lightgray}Fitzpatrick17k &skin type &vgg16 &0.09 $\pm$ 0.05\phantom{**} &0.16 $\pm$ 0.06*\phantom{*} &0.21 $\pm$ 0.08** &0.28 $\pm$ 0.02** \\
\bottomrule
\label{tab:TPR_gap_labelbias}
\end{tabular}
\end{table}

\begin{figure}
    \centering
    \includegraphics[width=1\linewidth]{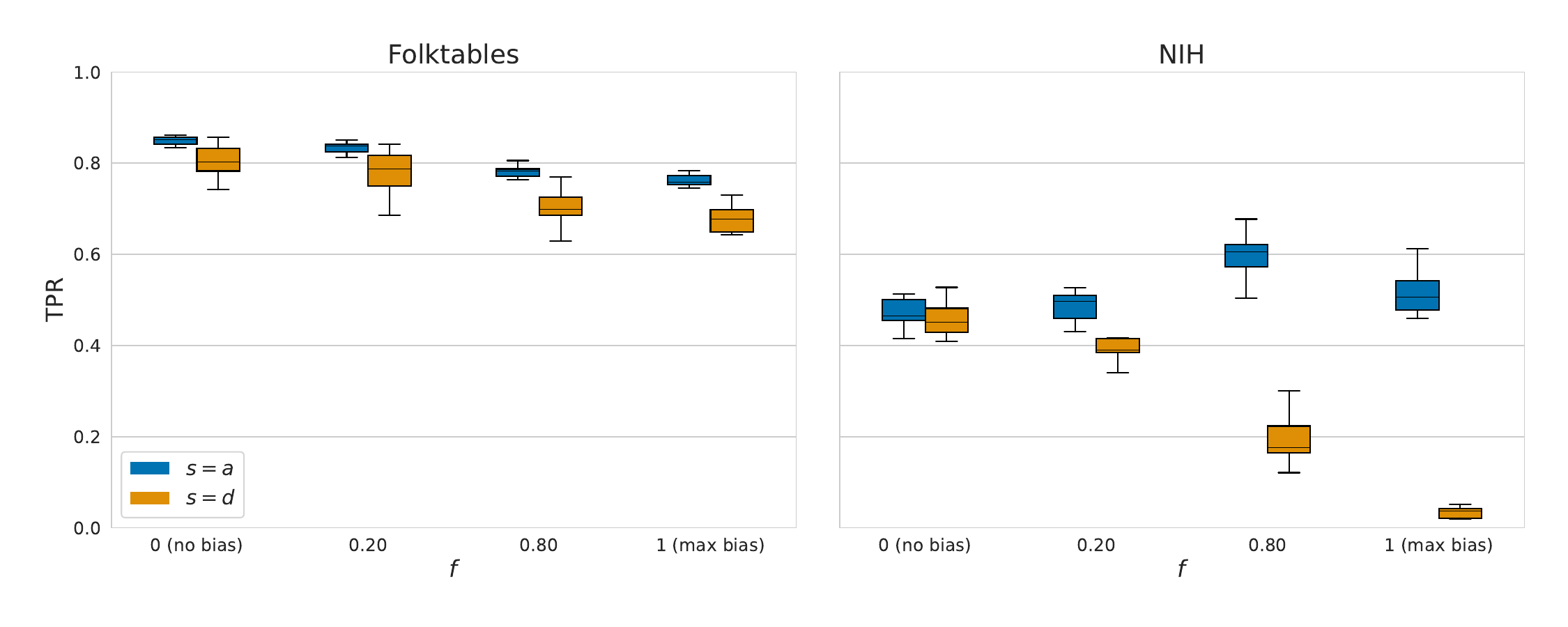}
    \caption{\textbf{Label bias induces sizable variations in groupwise true positive rates (TPR); the disadvantaged group is especially affected}. Boxplots representing the TPR on the advantaged and disadvantaged group ($y$ axis), as the percentage of disadvataged group items with flipped labels increases in the training set ($x$ axis).}
    \label{fig:TPR_gap_labelbias}
\end{figure}

\namedpar{Results} 
Table \ref{tab:TPR_gap_labelbias} shows that label bias has a large impact on fairness, sizably stronger than underrepresentation (Table \ref{tab:TPR_gap_underrepresentation}). Indeed, across all experiments, unfairness grows as $f$ increases. 
In datasets like NIH and Adult, this increase is very sizable, while for others such as Folktables it is more contained. Tables \ref{tab:pqp_f}, \ref{tab:eo_f}, and \ref{tab:dp_f} in \ref{app:exp_bias_detection} show a large increment in unfairness across the remaining metrics ($\text{PQP}$ and $\text{DP}$) and models (random forests and SVC).

Zooming in on this result, Figure \ref{fig:TPR_gap_labelbias} depicts the TPR for both sensitive groups on Folktables and NIH. We observe that label bias has a significant impact on the TPR of the disadvantaged group in both datasets, a trend observed consistently across all considered datasets.
On the other hand, the impact on the TPR of the advantaged group differs between datasets. Figure \ref{fig:TPR_gap_labelbias} shows a relatively stable TPR for NIH, while, for Folktables, the TPR of the advantaged group decreases with $f$.
Results for the remaining datasets are reported in Tables \ref{tab:tpr_d_f} and \ref{tab:tpr_a_f} in the appendix.

Broadly speaking, we distinguish two categories of datasets based on the effect of $f$ on the TPR of the advantaged group. Datasets such as Adult, Crime, Fitzpatrick17k, and NIH exhibit stable values for the TPR of the advantaged group while the TPR of the disadvantaged group decreases, therefore widening the gap. Conversely, datasets like German and Compas show patterns akin to Folktables, resulting in a less pronounced TPR gap. This diverging behavior is explained in Section~\ref{sec:proxies}. 

\begin{table}[!htp]\centering
\caption{\textbf{In the presence of weak label noise, it becomes preferable to omit the disadvantaged group from the training set}. The table below illustrates the Equal Opportunity difference ($\Delta \text{EO}$) between no representation and full representation for the disadvantaged group across two scenarios: one without label noise ($f=0$) and one with weak label noise $(f=0.2)$. Positive (negative) values indicate a relative improvement (decline) in the TPR of the disadvantaged group.}
\label{tab:deltaTPr20flip}
\scriptsize
\renewcommand{\arraystretch}{1.5}
\begin{tabular}{lllcc}
\toprule
& & & \multicolumn{2}{c}{$\Delta \text{EO}$}\\ \cmidrule(lr){4-5}
\textbf{Dataset} &\textbf{sensitive} &\textbf{model} &\multicolumn{1}{c}{$f=0$} & \multicolumn{1}{c}{$f=0.2$} \\ \midrule
\rowcolor{lightgray}&gender &
LR &\phantom{-}0.13 $\pm$ 0.07 &-0.01 $\pm$ 0.08 \\
\rowcolor{lightgray}\multirow{-2}{*}{Adult} &marital-status &
LR &-0.07 $\pm$ 0.05 &-0.13 $\pm$ 0.06 \\
 Compas &race &
LR &\phantom{-}0.01 $\pm$ 0.01 &0.02 $\pm$ 0.04 \\
\rowcolor{lightgray}Crime &race &
LR &-0.05 $\pm$ 0.17 &-0.11 $\pm$ 0.17 \\
Folktables &race &LR &\phantom{-}0.00 $\pm$ 0.06 &-0.01 $\pm$ 0.06  \\
\rowcolor{lightgray} German &age &
LR &-0.01 $\pm$ 0.15 &-0.09 $\pm$ 0.17  \\
 NIH &gender &DenseNet &\phantom{-}0.05 $\pm$ 0.03 &\phantom{-}0.03 $\pm$ 0.03 \\
\rowcolor{lightgray}Fitzpatrick17k &skin type &vgg16 &\phantom{-}0.02 $\pm$ 0.07 &-0.05 $\pm$ 0.08 \\
\bottomrule
\end{tabular}
\end{table}

\namedpar{On the joint effect of label bias and underrepresentation} Table \ref{tab:TPR_gap_labelbias} suggests that even a weak label bias can have a sizable impact on model fairness. We therefore study the joint effect of a weak label bias ($f \in \{0, 0.2\}$) and widely-ranging underrepresentation ($u \in \{0,1 \}$). Specifically, Table \ref{tab:deltaTPr20flip} summarizes the impact of excluding the disadvantaged group from the training set by presenting the difference between fairness under maximum underrepresentation ($\text{EO}_{u=1}$) and no underrepresentation ($\text{EO}_{u=0}$): 
\begin{align*}
    \Delta \text{EO} &= \text{EO}_{u=1} - \text{EO}_{u=0}.
\end{align*}
Positive values of $\Delta \text{EO}$ indicate that the inclusion of the disadvantaged group in the training set ($u=0$) leads to a decrease in the EO metric and, therefore, a relative improvement in their TPR. We quantify this improvement under no label bias ($f=0$) and weak label bias ($f=0.2$). As discussed in section \ref{sec:underrepresentation}, underrepresentation of the disadvantaged group in the training set (without label bias) has no clear effect on fairness, as confirmed by the first column of Table~\ref{tab:TPR_gap_labelbias} ($f=0$) displaying both positive and negative values. On the other hand, the second column ($f=0.2$) consistently displays negative values (with the exception of NIH and Compas). This means that, in the presence of relatively weak label bias, it may become preferable for the disadvantaged group to be \emph{completely omitted} from the training set.

Increasing the representation of the disadvantaged group in the training set under these conditions is not only unbeneficial but can, in some cases, be detrimental.

\namedpar{Interpretation}
The results presented in this section underscore the critical importance of precise and well-curated ground truth labels in datasets used for training classification models. Specifically, if the labels associated with one demographic group contain noise due to structural discrimination, this can significantly impact model fairness, thereby exacerbating existing biases. Our findings indicate that model performance for the disadvantaged group is consistently and substantially affected when the input labels for this group are subject to systematic bias. Conversely, label bias against the disadvantaged group has a weaker impact on the advantaged group, especially when proxies are strong (see Section~\ref{sec:proxies}); this discrepancy invariably leads to a fairness decline. Furthermore, we showed that even a small proportion of flipped labels can negatively affect the TPR gap. Overall, this shows that hastily adding disadvantaged groups into training sets without careful label curation can cause more harm than good for the members of those groups.

\subsection{Proxies}
\label{sec:proxies}

\begin{table}[!t]\centering
\caption{\textbf{The datasets most affected by label bias have strong proxies}. Strength of proxies for all datasets as measured by $s$AUC across 10 repetitions. We report sample means and standard deviations. The datasets where label bias leads to high unfairness, such as Adult, Crime and NIH (Table \ref{tab:TPR_gap_labelbias}), have high $s$AUC values.}
\label{tab:sensitiveAUC}
\scriptsize
\renewcommand{\arraystretch}{1.5} 
\begin{tabular}{llll}
\toprule
\textbf{Dataset} &\textbf{sensitive} &\textbf{model} &\textbf{sAUC} \\ \midrule
\rowcolor{lightgray}&gender & LR & 0.9349 $\pm$ 0.0025\\
\rowcolor{lightgray}\multirow{-2}{*}{Adult} &marital-status & LR & 0.9893 $\pm$ 0.0011\\
 Compas &race & LR & 0.6940 $\pm$ 0.0140\\
\rowcolor{lightgray}Crime &race & LR & 0.9847 $\pm$ 0.0074\\
Folktables &race &LR & 0.6821 $\pm$ 0.0112\\
\rowcolor{lightgray} German &age & LR & 0.7939 $\pm$ 0.0382\\
 NIH &gender &DenseNet & 0.9979 $\pm$ 0.0013\\
\rowcolor{lightgray}Fitzpatrick17k &skin type &vgg16 & 0.8946 $\pm$ 0.0100\\
\bottomrule
\end{tabular}
\end{table}

\namedpar{Setup} 
In this section, we study the effect of proxies on model fairness. We train classifiers $\hat{s}=h(x)$ to estimate the protected attribute $s$ and we compute their AUC to measure the strength of proxies ($s$AUC -- Equation \ref{eq:sauc_def}).\footnote{$h(x)$ takes as input the same features $x$ we used to train $g(x)$.}
To vary the strength of proxies, we leverage the subtractive protocol introduced in Section \ref{sec:bias_inject_method} by iteratively removing the feature that is most correlated with the protected attribute; we train a new classifier on the reduced input space and repeat this process until a random classifier performance is reached.
We study proxies in conjunction with label bias ($f \in (0,1)$).

\namedpar{Results}
Table~\ref{tab:sensitiveAUC} reports $s$AUC values for all datasets. Based on these values, we distinguish two types of datasets with diverging properties. Datasets like Adult (gender), Adult (marital-status), Crime, and NIH have very strong proxies for sensitive attributes ($s\text{AUC} > 0.9$), while Compas and Folktables have weaker proxies ($s\text{AUC} < 0.7$), indicating less information on sensitive attributes encoded in non-sensitive ones. This division mirrors Table \ref{tab:TPR_gap_labelbias}, where the four high-proxy datasets have the worst (highest) values for EO under maximum label bias ($f=1$) and three out of four — i.e. Adult (gender), Adult (marital-status), and NIH — already show statistically significant differences under low flip factors ($f=0.2$). Conversely, the negative effects of label bias are weaker for low-proxy datasets: both Folktables and Compas have no significant differences in EO even under maximum label bias $(f=1)$.

\begin{figure}
    \centering
    \begin{subfigure}[t]{0.48\textwidth}
    \includegraphics[width=1\linewidth]{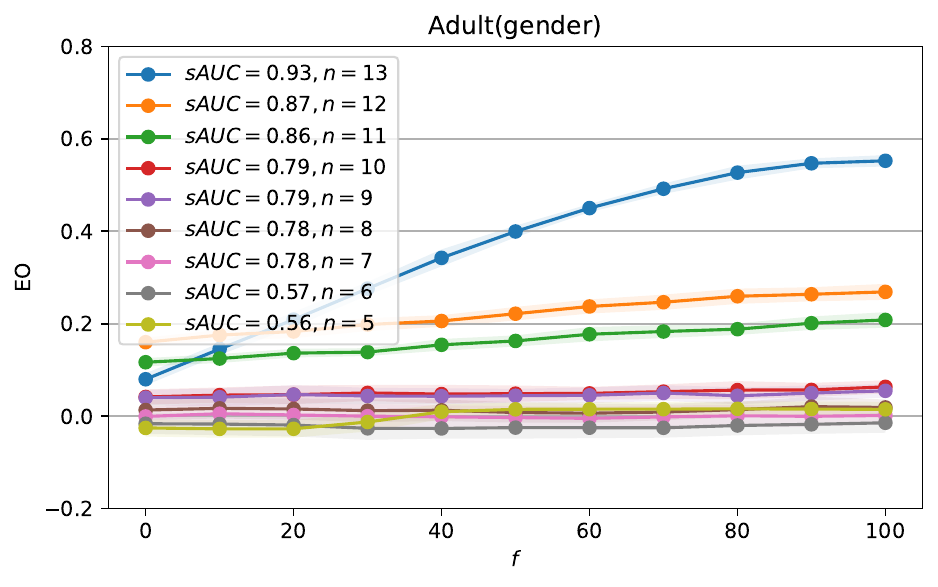}
    \label{fig:adultgender}
    \end{subfigure}
    ~
    \begin{subfigure}[t]{0.48\textwidth}
    \includegraphics[width=1\linewidth]{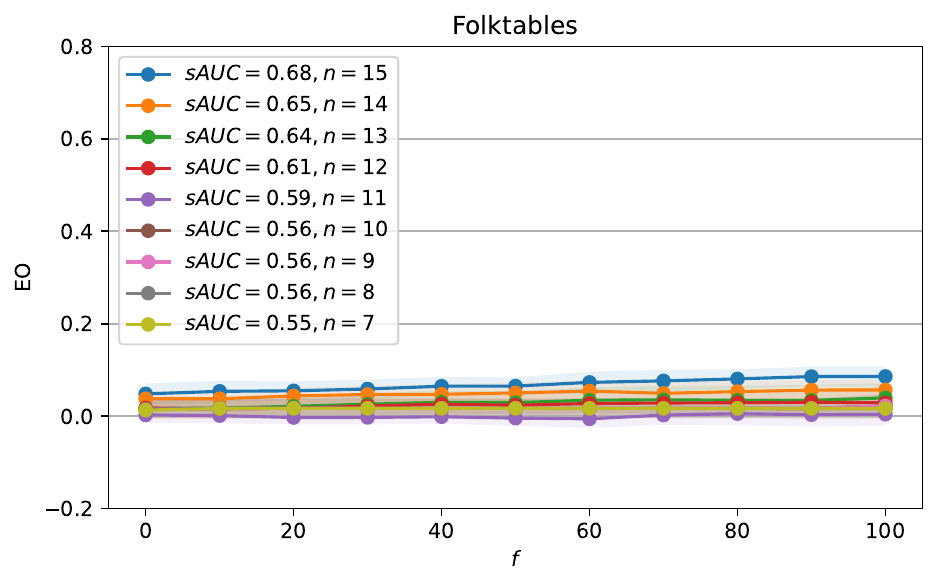}
    \label{fig:folktable}
    \end{subfigure}
    \caption{\textbf{Proxies exacerbate the risk of algorithmic discrimination caused by label bias}. EO ($y$ axis) increases with label bias $f$ ($x$ axis). This effect is mediated by proxies: weaker proxies (lower $s$AUC) correspond to a lower slope and a weaker effect of label bias on fairness.}\label{fig:proxiesFlip}
\end{figure}

To further investigate this trend, we jointly study label bias and proxies. Figure~\ref{fig:proxiesFlip} depicts EO for Adult (gender) and Folktables as the flip factor $f$ increases. Curves with different colors represent input spaces whose size (number of features $n$) is iteratively reduced by one. For both datasets, the impact of flips on EO (summarized by average curve slopes) decreases with $s$AUC. The decrease is more marked for Adult since it has stronger proxies and therefore starts from higher $s$AUC values, as reported in the legend.    

\namedpar{Interpretation}
These results prove that the presence of strong proxies amplifies the risk of algorithmic discrimination, particularly under label bias; when $s$AUC is large, the removal of correlated features can mitigate this risk by reducing the model's reliance on sensitive information. These findings highlight the critical role of proxy strength in exacerbating label bias and influencing the effectiveness of fairness interventions.

\section{Bias Detection}\label{sec:bias_detection}
In this section, we introduce mechanisms for bias detection. We evaluate their ability to highlight the presence of a specific type of bias in the data.

\subsection{Methods}
We propose three measures to detect the presence of each type of bias in the data. It is worth noting that in the previous section, we used a biased training set to train algorithms and an unbiased test set for their evaluation. In this section, we take the perspective of practitioners looking to evaluate their development dataset $\sigma$ for biases without necessarily having access to unbiased data sources. We therefore split $\sigma$ into identically distributed training and test sets. 

\namedpar{Underrepresentation}
We propose the Representation Difference (RD), to measure the underrepresentation of the disadvantaged group. 
%
\begin{equation}\label{eq:repratio}
    \text{RD}(\sigma) = \frac{|\sigma_a|-|\sigma_d|}{|\mathcal{\sigma}|} =  \Pr_{\sigma}(s=a) - \Pr_{\sigma}(s=d)
\end{equation}
RD quantifies the difference between the prevalence of the advantaged and disadvantaged groups. RD is a directional measure: positive (negative) values indicate a larger proportion of individuals from the (dis)advantaged group.  
A group $s$ can be considered fairly represented in $\sigma$ if $\text{RD}(\sigma)$ is within certain limits. For example, practitioners can pick a threshold based on the prevalence of $s=d$ in a target population.  

\namedpar{Label Bias}
We introduce two measures of systematic label noise against the disadvantaged group. Specifically, we train a base classifier $\hat{y}=g(x)$ and evaluate AUC curves distinguishing between sensitive groups. We let $p_g(x)$ denote the posterior probabilities obtained by the classifier and we define the cross-dataset AUC of $g(x)$ as
\begin{align*}
    \text{xAUC}_g(\sigma_1, \sigma_2) = \Pr(p_g(x_i) > p_g(x_j) | y_i=\oplus, y_j=\ominus, i \in \sigma_1, j \in \sigma_2),
\end{align*}
i.e. the probability that $g(x)$ correctly ranks a positive item from $\sigma_1$ higher than a negative one from $\sigma_2$.  

Our first measure, initially introduced by \cite{kallus2019fairnessriskscoresclassification}, leverages a partition of $\sigma$ into a disadvantaged set $\sigma_d$ and an advantaged set $\sigma_a$, computes both measures of cross-dataset AUC, and defines their difference as
\begin{align}
    \Delta \text{xAUC}_{\sigma} = \text{xAUC}(\sigma_a, \sigma_d) - \text{xAUC}(\sigma_d, \sigma_a). \label{eq:xauc}
\end{align}
Positive values indicate that pairs of advantaged positives and disadvantaged negatives are easier to separate correctly than pairs of disadvantaged positives and advantaged negatives. 

Next, we define the within-group AUC difference as 
\begin{align}
    \Delta \text{wAUC}_{\sigma} &= \text{AUC}_{\sigma_a}(g) - \text{AUC}_{\sigma_d}(g) \nonumber \\
    &= \Pr_{\sigma_a}(p_g(x_i) > p_g(x_j) \mid y_i=\oplus, y_j=\ominus) \nonumber \\
    &\quad - \Pr_{\sigma_d}(p_g(x_i) > p_g(x_j) \mid y_i=\oplus, y_j=\ominus), \label{eq:wauc}
\end{align}

i.e. we compute the AUC for advantaged ($s=a$) and disadvantaged items ($s=d$) separately, and measure their difference. Large absolute values indicate a better separability for one group. Notice that both Equations \eqref{eq:xauc} and \eqref{eq:wauc} are directional: positive (negative) values indicate better separability for the (dis)advantaged group.
Finally, we compute Separation Difference (SD) as their average
\begin{align}
    \text{SD}(\sigma) &= \frac{\Delta \text{xAUC}_{\sigma} + \Delta \text{wAUC}_{\sigma}}{2} \label{eq:dauc}
\end{align}
and employ it in the remainder of this section. We expect label bias to worsen the separability for the disadvantaged group and therefore yield high values of SD.

\namedpar{Proxies}
To quantify the information about protected features encoded in non-protected ones, which may act as proxies, we train a classifier $h(\cdot): x \rightarrow s$ to predict sensitive attributes from non-sensitive ones. The classifier's performance is then evaluated using the area under the ROC curve for the $s$ predictor ($s\text{AUC}$).
\begin{align}
    s\text{AUC} (\sigma) = \text{AUC}_{\sigma}(h)\label{eq:sauc}
\end{align}
Higher values of this metric indicate a better ability of the classifier to predict the sensitive feature from the non-sensitive one, indicating stronger proxies.

\subsection{Experiments}
\namedpar{Setup}
We leverage bias injection mechanisms to test bias detection.  We use part of $\sigma$ to train a classifier and part of $\sigma$ to evaluate its performance. We maintain an 80-10-10 train-validation-test split for tabular datasets and NIH and a 70-15-15 split for Fitzpatrick17k.
As in Section \ref{sec:3_experiments}, we subsample the disadvantaged group by varying the underrepresentation factor $u \in \{0, 0.2, 0.8, 1 \}$.
Similarly, we inject label bias letting the flip factor vary in $f \in \{0, 0.2, 0.8, 1 \}$.\footnote{Since extreme values such as $u=1$ and $f=1$ would render computation of SD and $s$AUC infeasible, we replace them with $u=0.95$ and $f=0.95$. It is worth reiterating that, in this section, training sets and test sets are drawn from the same distribution, differently from the previous section, where test sets were unbiased. \label{foot:instability}}
Finally, we inject proxies through the additive mechanism presented in Section \ref{sec:bias_inject_method}. We expand the input space with an additional feature derived as the sum of the sensitive attribute $s$ and a normal variable with zero mean and decreasing standard deviation.\footnote{For image datasets, the additional proxy is fed to the penultimate layer of the neural network.} The standard deviation varies to achieve Pearson correlation coefficients between the sensitive attribute and the additional feature of approximately $\{0, 0.25, 0.50, 0.75, 1\}$. A correlation of 0 corresponds to the baseline scenario with no additional feature, while a correlation of 1 reflects the maximally biased scenario in which the additional feature is identical to the protected attribute.

\begin{figure}[H]
    \centering
    \begin{subfigure}{\textwidth}
        \centering
        \includegraphics[width=\linewidth]{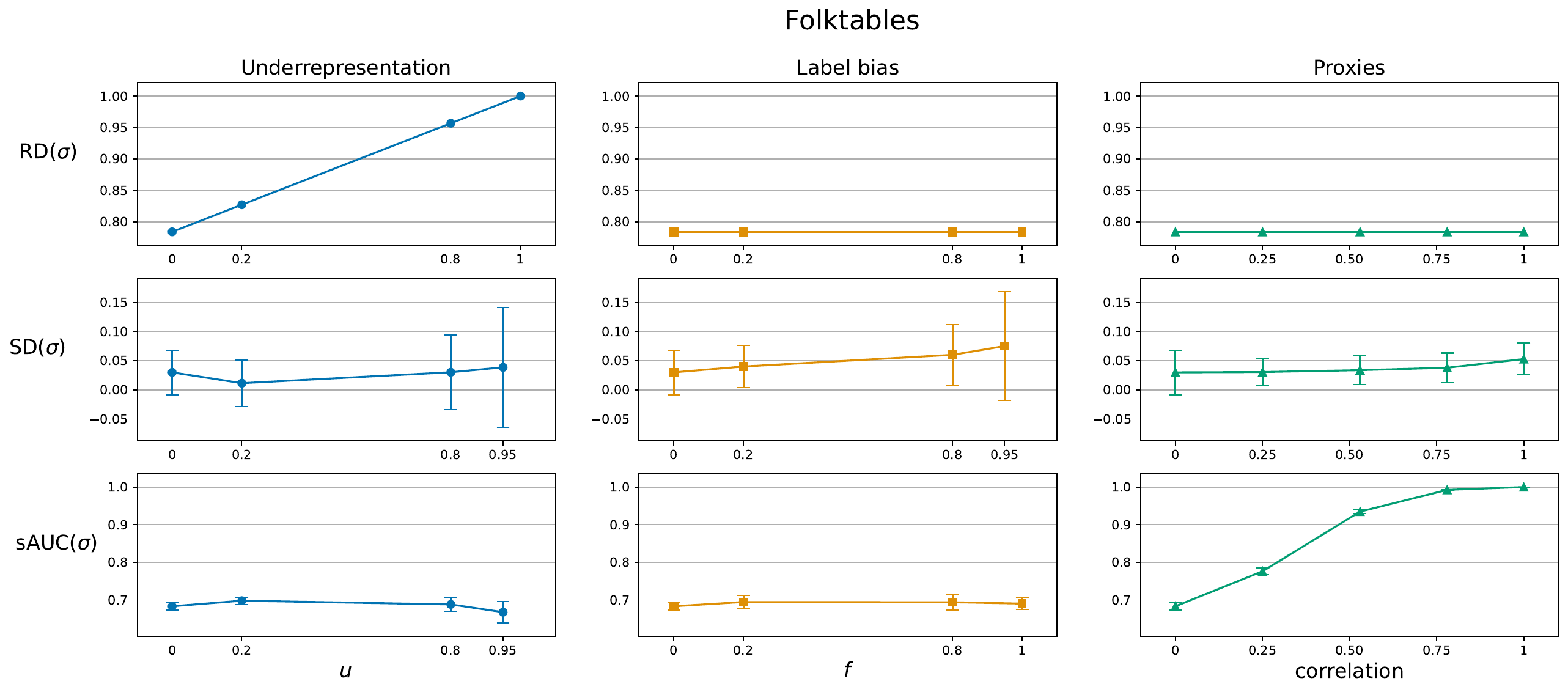}
    \end{subfigure}
    \begin{subfigure}{\textwidth}
        \centering
        \includegraphics[width=\linewidth]{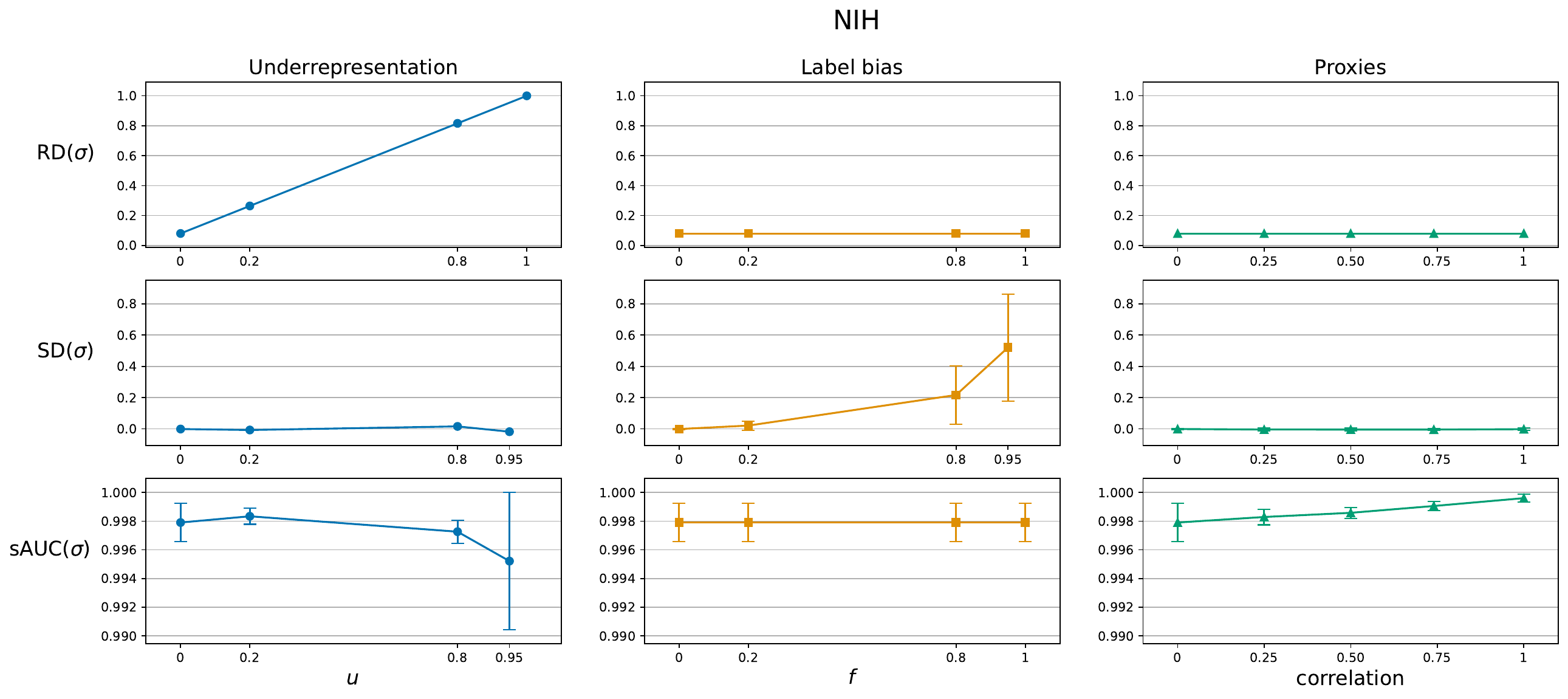}
    \end{subfigure}
    
    \caption{\textbf{The proposed measures capture specific types of bias}. Bias detection on Folktables and NIH. Columns correspond to three bias injection mechanisms; rows correspond to bias detection measures. Measures vary when the corresponding bias increases (diagonal) and remain relatively flat with other biases (off-diagonal).}
    
    \label{fig:bias_detection}
\end{figure}

\namedpar{Results}
In Figure \ref{fig:bias_detection}, we present bias detection results for Folktables and NIH. Experimental results for the other datasets are provided in \ref{app:exp_bias_detection}.
As anticipated, each bias detection metric specifically captures the type of bias it is designed to identify. The metrics exhibit strong variation in the diagonal panels of Figure \ref{fig:bias_detection}, while remaining relatively stable across the remaining panels. 

Specifically, underrepresentation is suitably captured by RD increasing linearly in the first column, while SD and $s$AUC exhibit minor oscillations in their average values. Notice that extreme underrepresentation leads to large variations around mean values for SD and $s$AUC due to numerical instability (see Footnote \ref{foot:instability}). In the second column, label bias leads to an increase in SD, while RD and $s$AUC remain constant. Finally, proxies of increasing strength are detected by $s$AUC in the third column, while RD and SD remain stable. 
These trends are consistent across nearly all datasets, except for the SD metric on Compas, whose increase with $f$ is barely noticeable.
Overall, these results show the effectiveness of the proposed measures in detecting specific data biases. 

\namedpar{Interpretation} 
After confirming the influence of data bias on algorithmic fairness in the previous section, in this section, we have provided a demonstration of bias detection based on principled measures. Each measure can detect a specific type of data bias. Crucially, their computation requires no access to unbiased test sets, making them widely applicable in practice.

\begin{figure}
    \centering
    \begin{subfigure}[H]{0.58\textwidth}
    \includegraphics[width=1\linewidth]{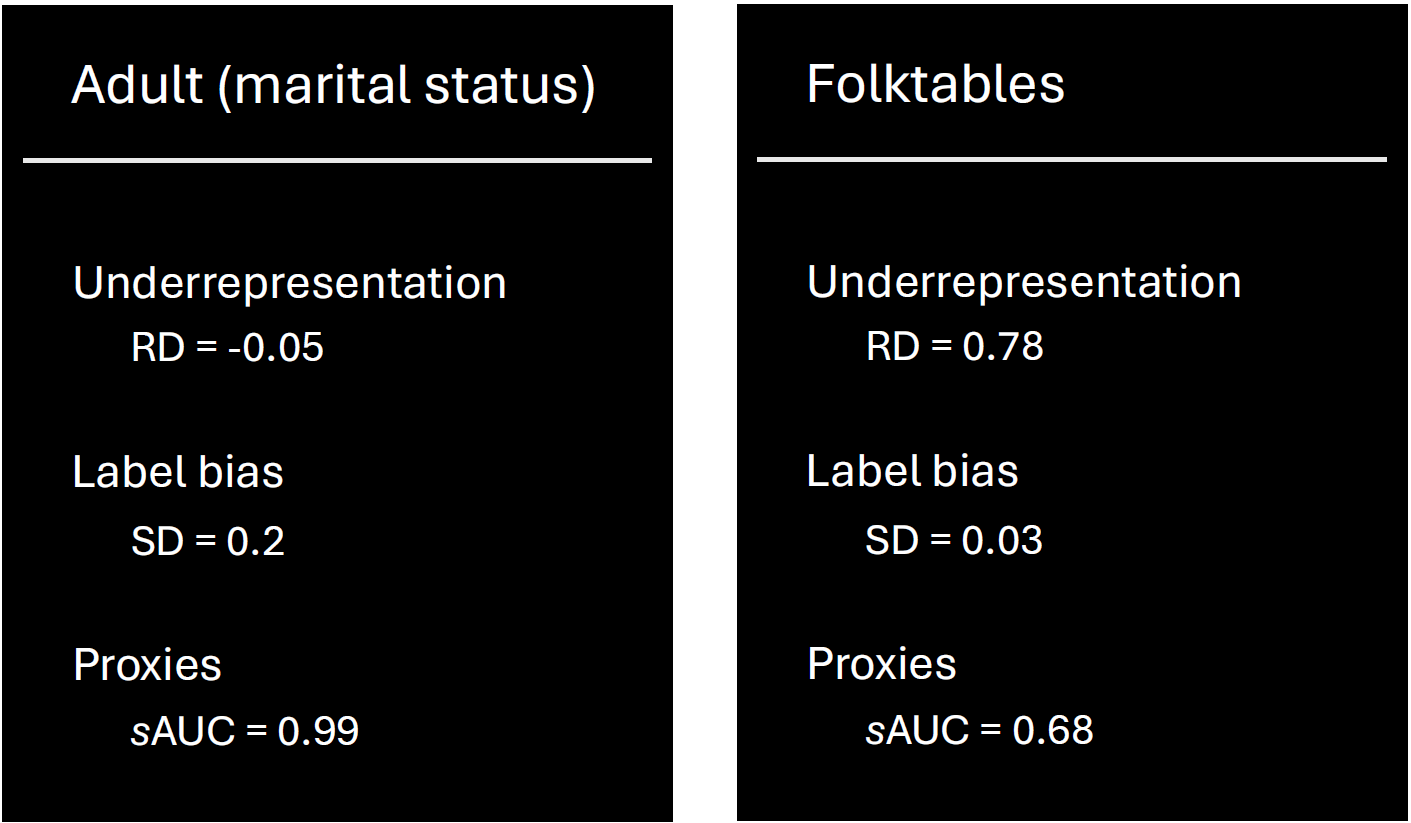}
    \end{subfigure}
    ~
    \begin{subfigure}[H]{0.363\textwidth}
    \includegraphics[width=1\linewidth]{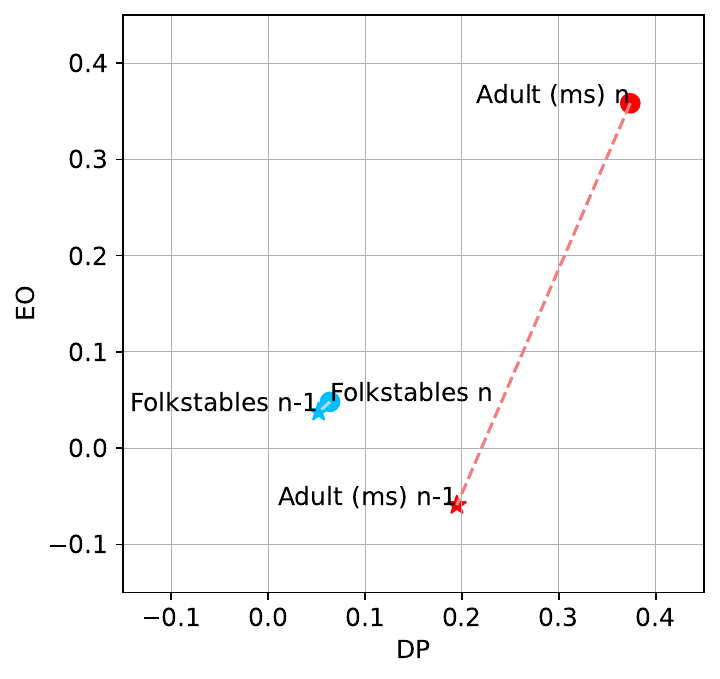}
    \end{subfigure}
    \caption{\textbf{Data Bias Profiles hint at the risk of algorithmic discrimination and effectiveness of fairness intervention}. DBP of Adult (left) and Folktables (center); on the right, model fairness summarized by demographic parity ($x$ axis) and equality of opportunity ($y$). DBP highlights strong proxies and label bias for Adult (marital status), leading to a high risk of discrimination (red round marker), which can be mitigated with proxy reduction (star-shaped marker). Folktables has weak proxies and label bias, translating into lower unfairness and ineffectiveness of proxy mitigation (blue markers).}\label{fig:dbp}
\end{figure}

\subsection{Data Bias Profile}
Based on these measures, we initiate the development of \emph{Data Bias Profiles} (DBP),
an extensible quantitative framework to describe data bias. 
We envision that the DBP will be used in fairness work to highlight biases that can lead to discrimination and inform decisions on fairness-enhancing interventions.
Additionally, DBP is suited to summarize biases in the documentation accompanying a dataset. 
We position this as an initial but foundational contribution—significant work remains to validate, refine, and extend the DBP into a mature, fully realized tool. Achieving this vision will require a coordinated effort across the research community to evolve the DBP into a robust and widely adopted quantitative framework.

Figure \ref{fig:dbp} demonstrates DBP with a practical use case. On the left, we present DBPs for two datasets. Adult (marital-status) presents considerable label bias and strong proxies. As shown in Section \ref{sec:3_experiments}, this entails a high probability of algorithmic discrimination. Folktables, on the other hand, has low label bias and weak proxies, highlighting a contained risk of algorithmic discrimination. This is confirmed by the right panel in Figure \ref{fig:dbp}, where round markers depict the average unfairness (DP and EO -- Equations \ref{eq:dp} and \ref{eq:eo}) for a logistic regression model over ten random splits of the datasets. DBPs also hint at the effectiveness of proxy reduction as a bias mitigation strategy. Tackling the strong proxies in Adult can largely reduce unfairness; Folktables, on the other hand, displays weak proxies and is unlikely to benefit from the same approach. We test this hypothesis by removing from each dataset the feature that is most strongly correlated with the sensitive attribute. Star-shaped markers depict EO and DP for the resulting models in the right panel of Figure \ref{fig:dbp}. As predicted, proxy removal strongly curbs unfairness on Adult (marital-status), while its effect on Folktables is barely noticeable.

\section{Discussion}
\label{sec:discussion}
We discuss our results in the broader context of responsible AI. Table \ref{tab:recommendation} summarizes the implications of this work for researchers and practitioners.

\namedpar{Underrepresentation in training is overemphasized}
Increasing the prevalence of vulnerable groups in training sets is touted as the key strategy to achieve fairness.
In stark contradiction with this credence, Section \ref{sec:3_experiments} shows that extreme variations in the prevalence of protected groups have a minor impact on fairness across diverse datasets, machine learning models, and metric choices. 
\emph{First and foremost, we urge practitioners and researchers against using these results as a blanket justification to neglect inclusion efforts.} Although challenging, expanding and diversifying datasets in a responsible manner is fundamental for keeping algorithmic systems in check. We provide a more nuanced interpretation. High-quality data from disadvantaged groups annotated with sensitive attributes is likely to be scarce, stemming e.g. from targeted curation efforts. Since including disadvantaged groups in training sets is often unlikely to bring meaningful improvements, we recommend prioritizing this data for reliable system evaluations (rather than training), including measurements of algorithmic fairness.  If fairness evaluations yield problematic results, practitioners should carry out an assessment of multiple bias factors that go beyond underrepresentation.

\begin{table}[t]
\caption{\textbf{Implications}. Takeaways and recommendations for algorithmic fairness researchers (R) and practitioners (P).}\label{tab:recommendation}
\scriptsize
\renewcommand{\arraystretch}{1.5}
\rowcolors{1}{white}{lightgray}
\begin{tabular}{m{5.5cm}m{7cm}}
\hiderowcolors
\toprule
\textbf{Takeaway} & \textbf{Recommendations}  \\
\hline
\showrowcolors
Underrepresentation in training is overemphasized &  \makecell[l]{(R) Critically reconsider widespread belief \\ (P) Use scarce annotated data for reliable evaluations} \\
Label bias is critical &  \makecell[l]{(R) Research techniques for label bias detection \\ (P) Seek and evaluate multiple target labels} \\
Data Bias Profiles (DBP) can link fairness \& bias &  \makecell[l]{(R) Diversify DBP in fairness testbeds  \\ (P) Use DBP to select fairness interventions \\ (P) Include DBP in data documentation}  \\
\bottomrule
\end{tabular}
\end{table}

\namedpar{Label bias is critical} 
Systematic bias against vulnerable groups in target variables (in short: label bias) is a common occurrence due to structural societal differences. Section \ref{sec:3_experiments} shows that label bias has a major impact on fairness across diverse metrics, models, and datasets as well as significant interactions with other types of data bias. For example, in the presence of label bias, increasing training set representation can have a detrimental effect on vulnerable groups. In this setting, underrepresentation may actually benefit vulnerable groups, challenging conventional wisdom. 
Notice that label bias is especially critical also because there is a high risk it will go undetected if models are trained and evaluated with datasets that exhibit the same bias (e.g. identically distributed training and test sets).
Based on these findings, we make two recommendations. Practitioners should seek multiple target variables for their models and carefully choose the most suitable one(s) to minimize the potential for label bias. This effort should blend qualitative approaches grounded in domain expertise with quantitative approaches for bias detection. In concert, researchers should develop reliable techniques for label bias detection. Section \ref{sec:bias_detection} is a first step in this direction. 

\namedpar{Data Bias Profiles (DBP) can link fairness and bias} 
Overall, this work contributes a list of independent data biases with mechanisms to study them (Section \ref{sec:3_databias}), a study of their joint influence on fairness (Section \ref{sec:3_experiments}), and principled methods for bias quantification (Section \ref{sec:bias_detection}). 
Building upon these contributions, we advocate a broader community effort to develop the Data Bias Profile (DBP), as a first attempt to summarize key bias indicators in a unified format. Rather than presenting a finalized framework, we offer an extensible prototype.
In their current form, DBPs are brief summaries of datasets 
that leverage bias quantification methods
for a principled analysis of fairness problems guided by data.
Model developers should use DBPs to reason about sources of model unfairness and select tailored approaches for mitigation. For example, detecting strong label bias may direct a developer towards fairness interventions for target label repair.
Additionally, DBPs can develop into reference documentation frameworks that practitioners will use to comply with data governance requirements, including bias detection provisions specified in the AI Act. 
From a research standpoint, the DBP offers promising directions.
DBPs can guide the development of fairness benchmarks, i.e. standardized collections to evaluate alternative fairness algorithms. For example, datasets can be distinguished based on the presence of strong proxies and weak proxies.
As we have shown, both vanilla algorithms (solely focused on accuracy) and fairness interventions behave differently based on the strength of proxies encoded in non-sensitive features.  
Fairness testbeds should thus include both strong-proxy and weak-proxy datasets to evaluate models under diverse conditions. 
Finally, DBPs may bridge hundreds of fairness algorithms \citep{hort2024bias} and datasets \citep{fabris2022algorithmic}, by helping to answer the key question: given a model that produces unfair predictions on a dataset, which type of fairness-enhancing algorithm is most suitable for that type of data and algorithm? 

\subsection*{Limitations}
This study has some limitations. 
First, we only cover three types of data bias. Although these are the most cited in scholarly articles and technical reports, different types of data bias are possible \citep{baumann2023bias,mehrabi2021survey}. Future work should consider additional biases, including feature bias, omitted variable bias, and concept drift across protected groups. 
Second, we consider binary protected attributes. While most results generalize to multi-group attributes by casting them as one-vs-all problems, this may become impractical for large cardinality $|\mathcal{S}|$. Natively catering to multi-group attributes will require careful adaptation.
Third, we provide no thresholds to distinguish between mild and excessive bias with our detection mechanisms. In its current form, the DBP is useful for relative comparisons across datasets; it will need further refinement to support thresholding. 
Fourth, while we experiment with popular and diverse fairness datasets, this is unlikely to be exhaustive of all settings encountered in practice. Future work should include additional datasets, with special attention to datasets with complementary properties. 
Finally, we note that it may be impossible to distinguish proper data bias, i.e. a shift between the data and a target population, from situations where the data is ``uncorrupted'', yet naturally encodes groupwise differences. Our work contributes a principled way to link numerical data properties with algorithmic fairness properties. Arguments on the source of those numerical properties are extremely valuable and complement our contributions.

\section{Conclusion}\label{sec:concl}
Data biases are key drivers of algorithmic discrimination. 
While this fact is broadly recognized, their relative importance and interaction remain understudied.
Our work targets this gap with a systematic study of bias conducive factors, their influence on algorithmic discrimination, and their detection through dedicated mechanisms.
These are necessary steps to develop a shared lexicon to describe data bias, document it unambiguously, and link it to fairness interventions in a principled fashion.
To realize these goals, we call for a community-wide effort to expand, formalize, and critically assess the Data Bias Profile, paving the way for a shared and trustworthy approach to quantitative bias documentation.

This line of work will be critical to steer anti-discrimination policy toward technically meaningful standards and to translate algorithmic fairness research into law-abiding practice.

\section*{Acknowledgements}
This work was supported by the Alexander von Humboldt Foun- dation, the FINDHR project (Horizon Europe grant agreement ID: 101070212) (A. Fabris) and by Ministero dell’Università della Ricerca (MUR),iniziativaDottoratiPON(M.Ceccon,G.A.Susto).

\clearpage
\appendix
\section{Datasets} \label{app:data}
In this appendix, we present algorithmic fairness datasets and their processing in this work. Sensitive features ($s$) that are used for fairness evaluations are excluded from input features.

\textbf{Adult}\footnote{\url{https://archive.ics.uci.edu/dataset/2/adult}} is a prominent dataset hosted by the UCI Machine Learning Repository, originally derived from the 1994 US Census database \citep{kohavi1996scaling}. The primary objective of this dataset is to predict whether an individual's annual income exceeds \$50,000. We follow \citet{donini2018empirical}, keeping all the features in the dataset. We use gender and marital-status as sensitive attributes.

\textbf{Compas}\footnote{\url{https://github.com/propublica/compas-analysis}} comprises data from the COMPAS (Correctional Offender Management Profiling for Alternative Sanctions) algorithm, a commercial tool used to predict recidivism among convicted individuals. The dataset, collected by ProPublica to audit the COMPAS system, surfaced discrimination against African-American defendants. We follow the preprocessing from \citet{ruoss2020learning} and use the following variables.
\begin{itemize}
    \item race ($s$): race of the defendants.
    \item age: age of the defendants.
    \item c\_charge\_degree: charge degree (F: Felony, M: Misdemeanor).
    \item diff\_custody: time spent in custody.
    \item diff\_jail: time spent in jail.
    \item sex: sex of the defendants.
    \item priors\_count: number of prior criminal records.
    \item length\_of\_stay: number of days spent in jail.
    \item v\_score\_text: COMPAS quantized score, summarizing additional features used by the model but unavailable in the data collected by ProPulica.
\end{itemize}
The target variable is two\_year\_recid, indicating whether an individual re-offended within 2 years of being released. For protected attributes, we focus on race, distinguishing between Caucasian (advantaged group) and African-American defendants (disadvantaged group).

\textbf{Crime}\footnote{\url{https://archive.ics.uci.edu/dataset/183/communities+and+crime}} is a real-world dataset from the UCI Machine Learning Repository, focused on predicting violent crime rates across various communities in the US. The task involves predicting whether a community can be classified as violent based on its crime rates, specifically when the number of crimes exceeds the median value of crimes across all states. We follow the setup from \cite{DBLP:conf/iclr/BalunovicRV22}, including binarized race as a sensitive attribute. We keep all the non-sensitive features for inference ($n=127$).

\textbf{Folktables}\footnote{\url{https://github.com/socialfoundations/folktables}} is a Python package designed to provide access to datasets derived from the US Census Bureau's American Community Survey (ACS) \citep{ding2021retiring}. The complete data underlying the folktables dataset comprises the full ACS census data, spanning all US states, multiple years, and prediction targets. In this work, we focus on the employment prediction task (ACSEmployment), filtering the data to include individuals aged between 16 and 90. We subsample at 1\% of the dataset size, stratifying on the target and sensitive label to maintain the distribution of the original data. We use the standard data loader keeping the following features:

\begin{itemize}
    \item ESR: employment status of the individual, represented as a binary categorical feature (1: Employed, 0: Otherwise). 
    \item RAC1P ($s$): detailed race recode (categorical values 1-9). 
    \item AGEP: age in years, with a maximum value of 99. 
    \item ANC: ancestry recode (categorical).  
    \item CIT: citizenship status of the individual, represented as a categorical string. 
    \item DEAR: hearing difficulty (binary).   
    \item DEYE: vision difficulty (binary).  
    \item DIS: disability recode (1: citizen with disability, 2: without). 
    \item DREM: cognitive difficulty of the individual, indicating if they have difficulty remembering, concentrating, or making decisions (binary).  
    \item ESP: employment status of parents (categorical)
    \item MAR: marital status of the individual (categorical). 
    \item MIG: Mobility status, indicating residence one year ago (categorical). 
    \item MIL: military service (categorical). 
     \item NATIVITY: binary variable indicating US native or foreign-born. 
    \item RELP: relationship (categorical values 1-17). 
    \item SCHL: educational attainment (categorical values 1-24, or NA). 
    \item SEX: sex/Gender (1: Male, 2: Female). 
\end{itemize}
The Employment Status Recode (ESR) is the target variable (equal to 1 if employed, 0 otherwise). 
The advantaged group consists of individuals with RAC1P equal to 1 (Caucasian), while the disadvantaged group includes all individuals with RAC1P values other than 1 (other races). 

\textbf{German\footnote{\url{https://archive.ics.uci.edu/dataset/522/south+german+credit}}} is another widely recognized dataset from the UCI Machine Learning Repository, encompassing records of bank loan applications in Germany. This dataset contains demographic and financial details of applicants, along with the loan approval outcomes. The primary prediction task is binary, aimed at determining creditworthiness based on loan repayment. We use the following features: 

\begin{itemize}
\item age ($s$): The age of the applicant, binarized with a 25-year threshold. 
\item amount: credit amount in Deutsche Mark. 
\item credit\_history: history of credit usage and repayment by the applicant (categorical). 
\item duration: duration of the loan in months (numeric). 
\item employment\_duration: tenure with current employer (numeric). 
\item housing: type of housing (categorical). 
\item installment\_rate: percentage of applicant's income allocated to loan installments (categorical). 
\item job: applicant's job and employability (categorical). 
 \item number\_credits: number of credits with this bank (categorical). 
\item other\_debtors: indication of an additional debtor or a guarantor for the credit (categorical). 
\item other\_installment\_plans: installment plans with other banks (categorical). 
\item people\_liable: number of people who are financially dependent on the applicant (categorical). 
\item property: applicant's most valuable property (categorical). 
\item purpose: purpose of loan (categorical). 
\item present\_residence: years lived at current address (categorical). 
\item savings: savings account balance (categorical). 
\item status: status of the individual's saving accounts (categorical). 
\item telephone: whether the applicant has a registered telephone line. 
\end{itemize}

\textbf{ChestX-ray14 (NIH)\footnote{\url{https://nihcc.app.box.com/v/ChestXray-NIHCC}}} is a comprehensive medical imaging dataset containing 112,120 frontal-view chest X-ray images from 30,805 unique patients, collected between 1992 and 2015 \citep{wang2017chestx}. Disease labels for fourteen common thoracic conditions were extracted from radiological reports using natural language processing. The labeled conditions include: atelectasis, cardiomegaly, consolidation, edema, effusion, emphysema, fibrosis, hernia, infiltration, mas, nodule, pneumonia, pneumothorax. The associated classification task is multi-label, with each of the 14 target labels indicating the presence of a specific disease. Additionally, patient metadata provides information on both gender and age.

We include only one image per patient, as previous studies have shown that this approach reduces bias without significantly affecting overall performance \citep{weng2023sexbased}. We conduct evaluations independently for each disease and report macro-averaged metrics across all diseases. In this study, binary gender is the sensitive attribute.

\textbf{Fitzpatrick17k\footnote{\url{https://github.com/mattgroh/fitzpatrick17k}}} is a medical imaging dataset containing 16,577 clinical images \citep{groh2021evaluating}, each annotated with labels for skin conditions and skin type based on the Fitzpatrick scale \citep{Fitzpatrick1988TheVA}. The images were sourced from two open-access dermatology atlases: 12,672 images from DermaAmin and 3,905 from Atlas Dermatologico.\footnote{\url{https://atlasdermatologico.com.br}} The dataset includes 114 distinct disease labels and two additional levels of aggregated skin condition classifications, structured according to the skin lesion taxonomy proposed by \cite{Esteva2017DermatologistlevelCO}. At the broadest classification level, skin conditions are divided into three main categories: benign lesions, malignant lesions and non-neoplastic lesions.
In a more detailed classification, skin conditions are categorized into nine types: inflammatory, malignant epidermal, genodermatoses, benign dermal, benign epidermal, malignant melanoma, benign melanocyte, malignant cutaneous lymphoma, malignant dermal. Our classification task is based on a binary variable derived from the highest-level skin condition classification, distinguishing between neoplastic, hence tumoral (either benign or malignant), and non-neoplastic diseases. This mimics a preliminary assessment for the presence of tumoral conditions through assistive technology used by dermatology experts. 
The Fitzpatrick skin type labels 
follow a six-point scale, with 1 being the lightest and 6 the darkest skin type. We binarize them into light (1-4) and very dark (5-6).


\FloatBarrier
\section{Additional Results on the Effect of Data Bias} \label{app:exp_effect_bias}
In this section, we report the effect of data bias on all machine learning models, datasets, and metrics. Specifically, performance is evaluated through balanced accuracy (Tables \ref{tab:bal_acc_a_u} and \ref{tab:bal_acc_a_f}), while fairness is assessed through prediction quality parity (PQP -- Tables \ref{tab:pqp_u} and \ref{tab:pqp_f}), equal opportunity (EO -- Tables \ref{tab:eo_u} and \ref{tab:eo_f}), and demographic parity (DP -- Tables \ref{tab:dp_u} and \ref{tab:dp_f}). We zoom in on EO by breaking it down into into groupwise true positive rate (TPR -- Tables \ref{tab:tpr_a_u}, \ref{tab:tpr_d_u}, \ref{tab:tpr_a_f}, and \ref{tab:tpr_d_f}) components. We provide results for underrepresentation (Appendix \ref{app:bias_u})  and label bias (Appendix \ref{app:bias_f}).

As in Section \ref{sec:3_experiments},  the tables include indicators of statistical significance; symbols (*) and (**) denote statistically significant differences with respect to the unbiased scenario thresholds at $p=0.05$ and $p=0.01$, respectively.

\subsection{Underrepresentation}\label{app:bias_u}

\begin{table}[H]\centering
\caption{Balanced accuracy varying the percentage of minority-group points retained in the training set from $u=0$ (no bias) to $u=1$ (maximum bias).}\label{tab:bal_acc_a_u}
\tiny
\renewcommand{\arraystretch}{1.5} 
\begin{tabular}{lllllll}\toprule
& & & \multicolumn{4}{c}{\textbf{Balanced Accuracy}}\\ \cmidrule(lr){4-7}
\textbf{Dataset} &\textbf{sensitive} &\textbf{model} &\multicolumn{1}{c}{$u=0$ \tiny{(no bias)}} & \multicolumn{1}{c}{$u=0.2$} & \multicolumn{1}{c}{$u=0.8$} & \multicolumn{1}{c}{$u=1$ \tiny{(max bias)}}\\ \midrule
\rowcolor{lightgray} & & LR & 0.77 $\pm$ 0.00 & 0.77 $\pm$ 0.00 & 0.77 $\pm$ 0.00 & 0.76 $\pm$ 0.01*\\
\rowcolor{lightgray} & & RF & 0.78 $\pm$ 0.00 & 0.78 $\pm$ 0.01 & 0.78 $\pm$ 0.00  & 0.77 $\pm$ 0.01*\\
\rowcolor{lightgray} & \multirow{-3}{*}{gender} & SVC & 0.77 $\pm$ 0.01 & 0.77 $\pm$ 0.01 & 0.77 $\pm$ 0.01 & 0.76 $\pm$ 0.01 \\
\rowcolor{lightgray} & & LR & 0.77 $\pm$ 0.00 & 0.77 $\pm$ 0.00 & 0.77 $\pm$ 0.00 & 0.77 $\pm$ 0.00\\
\rowcolor{lightgray} & & RF & 0.78 $\pm$ 0.00 & 0.78 $\pm$ 0.00 & 0.78 $\pm$ 0.00 & 0.78 $\pm$ 0.01 \\
\rowcolor{lightgray} \multirow{-6}{*}{Adult} &\multirow{-3}{*}{marital-status} & SVC & 0.77 $\pm$ 0.01 & 0.77 $\pm$ 0.01  & 0.77 $\pm$ 0.01 & 0.77 $\pm$ 0.01 \\
 & & LR & 0.67 $\pm$ 0.02 & 0.67 $\pm$ 0.02 & 0.67 $\pm$ 0.01 & 0.67 $\pm$ 0.01 \\
 & & RF & 0.69 $\pm$ 0.02 & 0.69 $\pm$ 0.01  & 0.69 $\pm$ 0.02  & 0.68 $\pm$ 0.02 \\
\multirow{-3}{*}{Compas} &\multirow{-3}{*}{race} & SVC & 0.65 $\pm$ 0.03 & 0.65 $\pm$ 0.03 & 0.65 $\pm$ 0.02  & 0.65 $\pm$ 0.01 \\
\rowcolor{lightgray} & & LR & 0.84 $\pm$ 0.02 & 0.84 $\pm$ 0.02 & 0.83 $\pm$ 0.02 & 0.83 $\pm$ 0.02 \\
\rowcolor{lightgray} & & RF & 0.83 $\pm$ 0.03 & 0.84 $\pm$ 0.03 & 0.83 $\pm$ 0.02 & 0.81 $\pm$ 0.03 \\
\rowcolor{lightgray}\multirow{-3}{*}{Crime} & \multirow{-3}{*}{race} & SVC & 0.84 $\pm$ 0.02 & 0.84 $\pm$ 0.02 & 0.83 $\pm$ 0.02 & 0.83 $\pm$ 0.03\\
 & & LR & 0.73 $\pm$ 0.01 & 0.73 $\pm$ 0.01 & 0.73 $\pm$ 0.01 &  0.73 $\pm$ 0.01\\
 & & RF & 0.78 $\pm$ 0.00 & 0.78 $\pm$ 0.01 & 0.77 $\pm$ 0.01* & 0.78 $\pm$ 0.00\\
\multirow{-3}{*}{Folktables} &\multirow{-3}{*}{race} & SVC & 0.72 $\pm$ 0.01 & 0.72 $\pm$ 0.01 & 0.72 $\pm$ 0.01 & 0.72 $\pm$ 0.01 \\
\rowcolor{lightgray} & & LR & 0.66 $\pm$ 0.06 & 0.65 $\pm$ 0.05 & 0.64 $\pm$ 0.07 & 0.65 $\pm$ 0.05 \\
\rowcolor{lightgray} & & RF & 0.68 $\pm$ 0.06 & 0.68 $\pm$ 0.05 & 0.65 $\pm$ 0.05 & 0.64 $\pm$ 0.05 \\
\rowcolor{lightgray} \multirow{-3}{*}{German} &\multirow{-3}{*}{age} & SVC & 0.66 $\pm$ 0.07 & 0.66 $\pm$ 0.07 & 0.63 $\pm$ 0.06 & 0.63 $\pm$ 0.06 \\
NIH &gender &DenseNet & 0.64 $\pm$ 0.01\phantom{**} & 0.64 $\pm$ 0.02\phantom{**} & 0.64 $\pm$ 0.02\phantom{**}& 0.63 $\pm$ 0.02\phantom{**}\\
\rowcolor{lightgray}Fitzpatrick17k &skin type &vgg16 & 0.73 $\pm$ 0.01\phantom{**}& 0.70 $\pm$ 0.01**& 0.73 $\pm$ 0.02\phantom{**}& 0.70 $\pm$ 0.02**\\
\bottomrule
\end{tabular}
\end{table}

\begin{table}[H]\centering
\caption{Prediction quality parity (PQP) varying the percentage of minority-group points retained in the training set from $u=0$ (no bias) to $u=1$ (maximum bias).}\label{tab:pqp_u}
\tiny
\renewcommand{\arraystretch}{1.5} 
\begin{tabular}{lllllll}\toprule
& & & \multicolumn{4}{c}{\textbf{PQP}}\\ \cmidrule(lr){4-7}
\textbf{Dataset} &\textbf{sensitive} &\textbf{model} &\multicolumn{1}{c}{$u=0$ \tiny{(no bias)}} & \multicolumn{1}{c}{$u=0.2$} & \multicolumn{1}{c}{$u=0.8$} & \multicolumn{1}{c}{$u=1$ \tiny{(max bias)}} \\ \midrule
\rowcolor{lightgray} & & LR & \phantom{-}0.00 $\pm$ 0.01\phantom{**} & \phantom{-}0.00 $\pm$ 0.01\phantom{**} & \phantom{-}0.01 $\pm$ 0.01\phantom{**} & \phantom{-}0.06 $\pm$ 0.03** \\
\rowcolor{lightgray} & & RF & \phantom{-}0.01 $\pm$ 0.01 & \phantom{-}0.01 $\pm$ 0.01 & \phantom{-}0.02 $\pm$ 0.01 & \phantom{-}0.11 $\pm$ 0.01** \\
\rowcolor{lightgray} &\multirow{-3}{*}{gender} & SVC & \phantom{-}0.01 $\pm$ 0.01 & \phantom{-}0.01 $\pm$ 0.01 & \phantom{-}0.01 $\pm$ 0.01 & \phantom{-}0.08 $\pm$ 0.01** \\
\rowcolor{lightgray} & & LR & \phantom{-}0.09 $\pm$ 0.01 & \phantom{-}0.09 $\pm$ 0.02 & \phantom{-}0.09 $\pm$ 0.02 & \phantom{-}0.06 $\pm$ 0.02** \\
\rowcolor{lightgray} & & RF & \phantom{-}0.08 $\pm$ 0.01 & \phantom{-}0.08 $\pm$ 0.01 & \phantom{-}0.09 $\pm$ 0.01 & \phantom{-}0.02 $\pm$ 0.02** \\
\rowcolor{lightgray} \multirow{-6}{*}{Adult} & \multirow{-3}{*}{marital-status} & SVC & \phantom{-}0.08 $\pm$ 0.02 & \phantom{-}0.09 $\pm$ 0.02 & \phantom{-}0.08 $\pm$ 0.02 & \phantom{-}0.06 $\pm$ 0.01* \\
 & & LR & -0.06 $\pm$ 0.03 & -0.06 $\pm$ 0.03 & -0.06 $\pm$ 0.04 & -0.06 $\pm$ 0.04 \\
 & & RF & -0.01 $\pm$ 0.05 & -0.02 $\pm$ 0.06 & -0.02 $\pm$ 0.05 & -0.02 $\pm$ 0.04 \\
\multirow{-3}{*}{Compas} &\multirow{-3}{*}{race} & SVC & -0.05 $\pm$ 0.03 & -0.05 $\pm$ 0.04 & -0.06 $\pm$ 0.03 & -0.06 $\pm$ 0.04 \\
\rowcolor{lightgray} & & LR & -0.02 $\pm$ 0.08 & -0.01 $\pm$ 0.09 & -0.04 $\pm$ 0.10 & -0.06 $\pm$ 0.10 \\
\rowcolor{lightgray} & & RF & -0.03 $\pm$ 0.05 & -0.04 $\pm$ 0.09 & -0.04 $\pm$ 0.07 & -0.08 $\pm$ 0.07 \\
\rowcolor{lightgray} \multirow{-3}{*}{Crime} &\multirow{-3}{*}{race} & SVC & -0.03 $\pm$ 0.08 & \phantom{-}0.01 $\pm$ 0.06 & -0.05 $\pm$ 0.05 & -0.06 $\pm$ 0.06 \\
& &LR & \phantom{-}0.01 $\pm$ 0.04 & \phantom{-}0.01 $\pm$ 0.04 & \phantom{-}0.02 $\pm$ 0.04 & \phantom{-}0.02 $\pm$ 0.04\\
& &RF & \phantom{-}0.01 $\pm$ 0.03 & \phantom{-}0.00 $\pm$ 0.03 & \phantom{-}0.01 $\pm$ 0.03 & \phantom{-}0.01 $\pm$ 0.03\\
\multirow{-3}{*}{Folktables} &\multirow{-3}{*}{race} &SVC & \phantom{-}0.02 $\pm$ 0.03 & \phantom{-}0.02 $\pm$ 0.03 & \phantom{-}0.02 $\pm$ 0.03 & \phantom{-}0.02 $\pm$ 0.03\\
\rowcolor{lightgray} & & LR & \phantom{-}0.07 $\pm$ 0.13 & \phantom{-}0.05 $\pm$ 0.09 & \phantom{-}0.08 $\pm$ 0.08 & \phantom{-}0.08 $\pm$ 0.10 \\
\rowcolor{lightgray} & & RF & -0.01 $\pm$ 0.07 & \phantom{-}0.02 $\pm$ 0.07 & \phantom{-}0.06 $\pm$ 0.05 & \phantom{-}0.04 $\pm$ 0.06 \\
\rowcolor{lightgray} \multirow{-3}{*}{German} &\multirow{-3}{*}{age} & SVC & \phantom{-}0.03 $\pm$ 0.10 & \phantom{-}0.03 $\pm$ 0.08 & \phantom{-}0.02 $\pm$ 0.10 & \phantom{-}0.02 $\pm$ 0.09 \\
NIH &gender &DenseNet & -0.01 $\pm$ 0.01 & \phantom{-}0.00 $\pm$ 0.01 & \phantom{-}0.01 $\pm$ 0.01** & \phantom{-}0.02 $\pm$ 0.01**\\
\rowcolor{lightgray}Fitzpatrick17k &skin type &vgg16 & -0.01 $\pm$ 0.01 & \phantom{-}0.03 $\pm$ 0.01** & \phantom{-}0.04 $\pm$ 0.02** & \phantom{-}0.03 $\pm$ 0.03**\\
\bottomrule
\end{tabular}
\end{table}

\begin{table}[H]\centering
\caption{True positive rate (TPR) of the advantaged group varying the percentage of minority-group points retained in the training set from $u=0$ (no bias) to $u=1$ (maximum bias).}\label{tab:tpr_a_u}
\tiny
\renewcommand{\arraystretch}{1.5} 
\begin{tabular}{lllllll}\toprule
& & & \multicolumn{4}{c}{\textbf{TPR ($s = a$)}}\\ \cmidrule(lr){4-7}
\textbf{Dataset} &\textbf{sensitive} &\textbf{model} &\multicolumn{1}{c}{$u=0$ \tiny{(no bias)}} & \multicolumn{1}{c}{$u=0.2$} & \multicolumn{1}{c}{$u=0.8$} & \multicolumn{1}{c}{$u=1$ \tiny{(max bias)}}\\ \midrule
\rowcolor{lightgray} & & LR & 0.61 $\pm$ 0.01\phantom{**} & 0.62 $\pm$ 0.01\phantom{**} & 0.62 $\pm$ 0.01\phantom{**} & 0.62 $\pm$ 0.01\phantom{**}\\
\rowcolor{lightgray} & & RF & 0.64 $\pm$ 0.01 & 0.63 $\pm$ 0.01 & 0.64 $\pm$ 0.01 & 0.63 $\pm$ 0.01\\
\rowcolor{lightgray} &\multirow{-3}{*}{gender} & SVC & 0.61 $\pm$ 0.01 & 0.61 $\pm$ 0.01 & 0.61 $\pm$ 0.01 & 0.61 $\pm$ 0.01\\
\rowcolor{lightgray} & & LR & 0.65 $\pm$ 0.01 & 0.65 $\pm$ 0.01 & 0.65 $\pm$ 0.01 & 0.66 $\pm$ 0.01 \\
\rowcolor{lightgray} & & RF & 0.66 $\pm$ 0.01 & 0.66 $\pm$ 0.01 & 0.66 $\pm$ 0.01 & 0.67 $\pm$ 0.01 \\
\rowcolor{lightgray} \multirow{-6}{*}{Adult} &\multirow{-3}{*}{marital-status} & SVC & 0.64 $\pm$ 0.01 & 0.64 $\pm$ 0.01 & 0.65 $\pm$ 0.01 & 0.66 $\pm$ 0.01** \\
  & & LR & 0.85 $\pm$ 0.02  & 0.86 $\pm$ 0.01  & 0.87 $\pm$ 0.02  & 0.88 $\pm$ 0.02* \\
  & & RF & 0.81 $\pm$ 0.03  & 0.81 $\pm$ 0.03  & 0.81 $\pm$ 0.02  & 0.81 $\pm$ 0.03  \\
 \multirow{-3}{*}{Compas} &\multirow{-3}{*}{race} & SVC & 0.85 $\pm$ 0.04  & 0.86 $\pm$ 0.05  & 0.91 $\pm$ 0.02**  & 0.92 $\pm$ 0.02** \\
\rowcolor{lightgray} & & LR & 0.90 $\pm$ 0.04 & 0.90 $\pm$ 0.04 & 0.89 $\pm$ 0.04 & 0.91 $\pm$ 0.04 \\
\rowcolor{lightgray} & & RF & 0.90 $\pm$ 0.03 & 0.90 $\pm$ 0.03 & 0.91 $\pm$ 0.03 & 0.92 $\pm$ 0.03 \\
\rowcolor{lightgray}\multirow{-3}{*}{Crime} &\multirow{-3}{*}{race} & SVC & 0.91 $\pm$ 0.04 & 0.91 $\pm$ 0.03 & 0.91 $\pm$ 0.03 & 0.93 $\pm$ 0.03 \\
 & &LR & 0.83 $\pm$ 0.01 & 0.83 $\pm$ 0.01 & 0.83 $\pm$ 0.01 & 0.83 $\pm$ 0.01\\
 & &RF & 0.85 $\pm$ 0.01 & 0.86 $\pm$ 0.01 & 0.86 $\pm$ 0.01 & 0.86 $\pm$ 0.01\\
\multirow{-3}{*}{Folktables} &\multirow{-3}{*}{race} &SVC & 0.85 $\pm$ 0.01 & 0.85 $\pm$ 0.01 & 0.85 $\pm$ 0.01 & 0.85 $\pm$ 0.01\\
\rowcolor{lightgray} & & LR & 0.89 $\pm$ 0.05 & 0.89 $\pm$ 0.05 & 0.91 $\pm$ 0.05 & 0.91 $\pm$ 0.05 \\
\rowcolor{lightgray} & & RF & 0.93 $\pm$ 0.02 & 0.91 $\pm$ 0.04 & 0.93 $\pm$ 0.02 & 0.92 $\pm$ 0.06 \\
\rowcolor{lightgray} \multirow{-3}{*}{German} &\multirow{-3}{*}{age} & SVC & 0.93 $\pm$ 0.02 & 0.92 $\pm$ 0.02 & 0.93 $\pm$ 0.04 & 0.94 $\pm$ 0.04 \\
NIH &gender &DenseNet & 0.46 $\pm$ 0.03 & 0.49 $\pm$ 0.02 & 0.45 $\pm$ 0.05 &  0.45 $\pm$ 0.04\\
\rowcolor{lightgray} Fitzpatrick17k &skin type &vgg16 & 0.70 $\pm$ 0.02 & 0.68 $\pm$ 0.02 & 0.68 $\pm$ 0.01* & 0.69 $\pm$ 0.01\\
\bottomrule
\end{tabular}
\end{table}

\begin{table}[H]\centering
\caption{True positive rate (TPR) of the disadvantaged group varying the percentage of minority-group points retained in the training set from $u=0$ (no bias) to $u=1$ (maximum bias).}\label{tab:tpr_d_u}
\tiny
\renewcommand{\arraystretch}{1.5} 
\begin{tabular}{lllllll}\toprule
& & & \multicolumn{4}{c}{\textbf{TPR ($s=d$)}}\\ \cmidrule(lr){4-7}
\textbf{Dataset} &\textbf{sensitive} &\textbf{model} &\multicolumn{1}{c}{$u=0$ \tiny{(no bias)}} & \multicolumn{1}{c}{$u=0.2$} & \multicolumn{1}{c}{$u=0.8$} & \multicolumn{1}{c}{$u=1$ \tiny{(max bias)}}\\ \midrule
\rowcolor{lightgray} & & LR & 0.53 $\pm$ 0.02\phantom{**} & 0.53 $\pm$ 0.02\phantom{**} & 0.53 $\pm$ 0.02\phantom{**} & 0.41 $\pm$ 0.07**\\
\rowcolor{lightgray} & & RF & 0.54 $\pm$ 0.02 & 0.54 $\pm$ 0.02 & 0.52 $\pm$ 0.02 & 0.34 $\pm$ 0.03**\\
\rowcolor{lightgray} &\multirow{-3}{*}{gender} & SVC & 0.53 $\pm$ 0.02 & 0.53 $\pm$ 0.02 & 0.51 $\pm$ 0.03 & 0.36 $\pm$ 0.04**\\
\rowcolor{lightgray} & & LR & 0.29 $\pm$ 0.03 & 0.29 $\pm$ 0.03 & 0.29 $\pm$ 0.03 & 0.37 $\pm$ 0.03** \\
\rowcolor{lightgray} & & RF & 0.35 $\pm$ 0.03 & 0.34 $\pm$ 0.03 & 0.33 $\pm$ 0.02 & 0.51 $\pm$ 0.04** \\
\rowcolor{lightgray} \multirow{-6}{*}{Adult} &\multirow{-3}{*}{marital-status} & SVC & 0.30 $\pm$ 0.03 & 0.30 $\pm$ 0.03 & 0.31 $\pm$ 0.03 & 0.39 $\pm$ 0.02**\\
 & & LR & 0.68 $\pm$ 0.03 & 0.68 $\pm$ 0.03 & 0.71 $\pm$ 0.03 & 0.72 $\pm$ 0.02**\\
 & & RF & 0.66 $\pm$ 0.03 & 0.67 $\pm$ 0.03 & 0.68 $\pm$ 0.04 & 0.70 $\pm$ 0.03** \\
 \multirow{-3}{*}{Compas} &\multirow{-3}{*}{race} & SVC & 0.68 $\pm$ 0.06 & 0.69 $\pm$ 0.07 & 0.76 $\pm$ 0.05* & 0.78 $\pm$ 0.03** \\
 \rowcolor{lightgray} & & LR & 0.57 $\pm$ 0.10 & 0.54 $\pm$ 0.12 & 0.59 $\pm$ 0.14 & 0.63 $\pm$ 0.13 \\
 \rowcolor{lightgray} & & RF & 0.57 $\pm$ 0.07 & 0.58 $\pm$ 0.10 & 0.62 $\pm$ 0.08 & 0.75 $\pm$ 0.09** \\
\rowcolor{lightgray}\multirow{-3}{*}{Crime} &\multirow{-3}{*}{race} & SVC & 0.59 $\pm$ 0.08 & 0.55 $\pm$ 0.08 & 0.61 $\pm$ 0.07 & 0.63 $\pm$ 0.09 \\
 & & LR & 0.79 $\pm$ 0.04 & 0.79 $\pm$ 0.04 & 0.78 $\pm$ 0.04 & 0.78 $\pm$ 0.05\\
 & & RF & 0.83 $\pm$ 0.03 & 0.84 $\pm$ 0.04 & 0.85 $\pm$ 0.03 & 0.84 $\pm$ 0.04\\
\multirow{-3}{*}{Folktables} &\multirow{-3}{*}{race} &SVC & 0.80 $\pm$ 0.04 & 0.81 $\pm$ 0.04 & 0.81 $\pm$ 0.04 & 0.81 $\pm$ 0.03\\
\rowcolor{lightgray} & & LR & 0.82 $\pm$ 0.13 & 0.82 $\pm$ 0.12 & 0.84 $\pm$ 0.10 & 0.85 $\pm$ 0.09 \\
\rowcolor{lightgray} & & RF & 0.90 $\pm$ 0.07 & 0.87 $\pm$ 0.07 & 0.90 $\pm$ 0.07 & 0.89 $\pm$ 0.07 \\
\rowcolor{lightgray} \multirow{-3}{*}{German} &\multirow{-3}{*}{age} & SVC & 0.87 $\pm$ 0.09 & 0.89 $\pm$ 0.09 & 0.92 $\pm$ 0.10 & 0.90 $\pm$ 0.11 \\
NIH &gender &DenseNet & 0.44 $\pm$ 0.05 & 0.45 $\pm$ 0.03 & 0.42 $\pm$ 0.04 & 0.39 $\pm$ 0.03* \\
\rowcolor{lightgray}Fitzpatrick17k &skin type &vgg16 & 0.61 $\pm$ 0.05 & 0.58 $\pm$ 0.06 & 0.57 $\pm$ 0.05 & 0.58 $\pm$ 0.05\\
\bottomrule
\end{tabular}
\end{table}

\begin{table}[H]\centering
\caption{Equal opportunity (EO) varying the percentage of minority-group points retained in the training set from $u=0$ (no bias) to $u=1$ (maximum bias).}\label{tab:eo_u}
\tiny
\renewcommand{\arraystretch}{1.5} 
\begin{tabular}{lllllll}\toprule
& & & \multicolumn{4}{c}{\textbf{EO}}\\ \cmidrule(lr){4-7}
\textbf{Dataset} &\textbf{sensitive} &\textbf{model} &\multicolumn{1}{c}{$u=0$ \tiny{(no bias)}} & \multicolumn{1}{c}{$u=0.2$} & \multicolumn{1}{c}{$u=0.8$} & \multicolumn{1}{c}{$u=1$ \tiny{(max bias)}}\\ \midrule
\rowcolor{lightgray} & & LR & 0.08 $\pm$ 0.02 & 0.09 $\pm$ 0.02 & 0.09 $\pm$ 0.02 & 0.21 $\pm$ 0.07**\\
\rowcolor{lightgray} & & RF & 0.10 $\pm$ 0.02 & 0.09 $\pm$ 0.02 & 0.12 $\pm$ 0.03 & 0.30 $\pm$ 0.03**\\
\rowcolor{lightgray} &\multirow{-3}{*}{gender} & SVC & 0.08 $\pm$ 0.02 & 0.08 $\pm$ 0.02 & 0.10 $\pm$ 0.03 & 0.25 $\pm$ 0.03**\\
\rowcolor{lightgray} & & LR & 0.36 $\pm$ 0.03 & 0.36 $\pm$ 0.03 & 0.37 $\pm$ 0.03 & 0.29 $\pm$ 0.04** \\
\rowcolor{lightgray} & & RF & 0.31 $\pm$ 0.03 & 0.31 $\pm$ 0.03 & 0.33 $\pm$ 0.03 & 0.16 $\pm$ 0.05** \\
\rowcolor{lightgray} \multirow{-6}{*}{Adult} &\multirow{-3}{*}{marital-status} & SVC & 0.34 $\pm$ 0.03 & 0.34 $\pm$ 0.03 & 0.34 $\pm$ 0.03 & 0.27 $\pm$ 0.02**\\
 & & LR & 0.17 $\pm$ 0.03 & 0.17 $\pm$ 0.03 & 0.16 $\pm$ 0.02 & 0.16 $\pm$ 0.02 \\
 & & RF & 0.15 $\pm$ 0.05 & 0.14 $\pm$ 0.05 & 0.13 $\pm$ 0.04 & 0.11 $\pm$ 0.05 \\
 \multirow{-3}{*}{Compas} &\multirow{-3}{*}{race} & SVC & 0.18 $\pm$ 0.04 & 0.17 $\pm$ 0.04 & 0.15 $\pm$ 0.04 & 0.13 $\pm$ 0.03** \\
 \rowcolor{lightgray} & & LR & 0.33 $\pm$ 0.11 & 0.36 $\pm$ 0.12 & 0.30 $\pm$ 0.14 & 0.28 $\pm$ 0.13 \\
 \rowcolor{lightgray} & & RF & 0.33 $\pm$ 0.06 & 0.32 $\pm$ 0.11 & 0.29 $\pm$ 0.08 & 0.16 $\pm$ 0.09**\\
\rowcolor{lightgray}\multirow{-3}{*}{Crime} &\multirow{-3}{*}{race} & SVC & 0.32 $\pm$ 0.08 & 0.36 $\pm$ 0.09 & 0.30 $\pm$ 0.07 & 0.31 $\pm$ 0.07 \\
 & & LR & 0.05 $\pm$ 0.04 & 0.05 $\pm$ 0.04 & 0.05 $\pm$ 0.04 & 0.05 $\pm$ 0.04\\
 & & RF & 0.02 $\pm$ 0.04 & 0.02 $\pm$ 0.03 & 0.02 $\pm$ 0.03 & 0.02 $\pm$ 0.04\\
\multirow{-3}{*}{Folktables} &\multirow{-3}{*}{race} & SVC & 0.04 $\pm$ 0.03 & 0.04 $\pm$ 0.04 & 0.04 $\pm$ 0.03 & 0.04 $\pm$ 0.03\\
\rowcolor{lightgray} & & LR & 0.07 $\pm$ 0.13 & 0.06 $\pm$ 0.11 & 0.07 $\pm$ 0.09 & 0.06 $\pm$ 0.07 \\
\rowcolor{lightgray} & & RF & 0.03 $\pm$ 0.06 & 0.03 $\pm$ 0.06 & 0.03 $\pm$ 0.07 & 0.03 $\pm$ 0.08 \\
\rowcolor{lightgray} \multirow{-3}{*}{German} &\multirow{-3}{*}{age} & SVC & 0.06 $\pm$ 0.08 & 0.03 $\pm$ 0.08 & 0.01 $\pm$ 0.08 & 0.04 $\pm$ 0.08 \\
NIH & gender & DenseNet & 0.01 $\pm$ 0.02 & 0.04 $\pm$ 0.01** & 0.03 $\pm$ 0.02 & 0.06 $\pm$ 0.02**\\
\rowcolor{lightgray} Fitzpatrick17k & skin type & vgg16 & 0.09 $\pm$ 0.05 & 0.11 $\pm$ 0.06 & 0.11 $\pm$ 0.05 & 0.11 $\pm$ 0.05\\
\bottomrule
\end{tabular}
\end{table}

\begin{table}[H]\centering
\caption{Demographic parity (DP) varying the percentage of minority-group points retained in the training set from $u=0$ (no bias) to $u=1$ (maximum bias).}\label{tab:dp_u}
\tiny
\renewcommand{\arraystretch}{1.5} 
\begin{tabular}{lllllll}\toprule
& & & \multicolumn{4}{c}{\textbf{DP}}\\ \cmidrule(lr){4-7}
\textbf{Dataset} &\textbf{sensitive} &\textbf{model} &\multicolumn{1}{c}{$u=0$ \tiny{(no bias)}} & \multicolumn{1}{c}{$u=0.2$} & \multicolumn{1}{c}{$u=0.8$} & \multicolumn{1}{c}{$u=1$ \tiny{(max bias)}} \\ \midrule
\rowcolor{lightgray} & & LR & 0.18 $\pm$ 0.01\phantom{**} & 0.18 $\pm$ 0.01\phantom{**} & 0.18 $\pm$ 0.01\phantom{**} & 0.20 $\pm$ 0.01** \\
\rowcolor{lightgray} & & RF & 0.18 $\pm$ 0.01 & 0.18 $\pm$ 0.01 & 0.18 $\pm$ 0.01 & 0.21 $\pm$ 0.01** \\
\rowcolor{lightgray} &\multirow{-3}{*}{gender} & SVC & 0.17 $\pm$ 0.00 & 0.17 $\pm$ 0.00 & 0.18 $\pm$ 0.01* & 0.20 $\pm$ 0.01** \\
\rowcolor{lightgray}  & & LR & 0.37 $\pm$ 0.01 & 0.37 $\pm$ 0.01 & 0.38 $\pm$ 0.01 & 0.36 $\pm$ 0.02 \\
\rowcolor{lightgray}  & & RF & 0.36 $\pm$ 0.02 & 0.35 $\pm$ 0.02 & 0.36 $\pm$ 0.01 & 0.31 $\pm$ 0.02** \\
\rowcolor{lightgray} \multirow{-6}{*}{Adult} &\multirow{-3}{*}{marital-status} & SVC & 0.36 $\pm$ 0.01 & 0.36 $\pm$ 0.01 & 0.36 $\pm$ 0.01 & 0.34 $\pm$ 0.01** \\
 & & LR & 0.26 $\pm$ 0.03 & 0.27 $\pm$ 0.03 & 0.26 $\pm$ 0.03 & 0.25 $\pm$ 0.03 \\
 & & RF & 0.21 $\pm$ 0.05 & 0.20 $\pm$ 0.04 & 0.19 $\pm$ 0.04 & 0.17 $\pm$ 0.05 \\
\multirow{-3}{*}{Compas} &\multirow{-3}{*}{race} & SVC & 0.26 $\pm$ 0.02 & 0.25 $\pm$ 0.02 & 0.23 $\pm$ 0.03** & 0.22 $\pm$ 0.03** \\
\rowcolor{lightgray} & & LR & 0.62 $\pm$ 0.06 & 0.63 $\pm$ 0.05 & 0.61 $\pm$ 0.05 & 0.61 $\pm$ 0.05 \\
\rowcolor{lightgray} & & RF & 0.62 $\pm$ 0.07 & 0.63 $\pm$ 0.07 & 0.61 $\pm$ 0.06 & 0.53 $\pm$ 0.07** \\
\rowcolor{lightgray} \multirow{-3}{*}{Crime} &\multirow{-3}{*}{race} & SVC & 0.62 $\pm$ 0.07 & 0.62 $\pm$ 0.06 & 0.61 $\pm$ 0.05 & 0.63 $\pm$ 0.06 \\
 & &LR & 0.06 $\pm$ 0.03 & 0.06 $\pm$ 0.03 & 0.06 $\pm$ 0.03 & 0.06 $\pm$ 0.03\\
 & &RF & 0.05 $\pm$ 0.03 & 0.05 $\pm$ 0.03 & 0.04 $\pm$ 0.04 & 0.04 $\pm$ 0.03\\
\multirow{-3}{*}{Folktables} &\multirow{-3}{*}{race} &SVC & 0.06 $\pm$ 0.03 & 0.06 $\pm$ 0.03 & 0.05 $\pm$ 0.02 & 0.05 $\pm$ 0.02\\
\rowcolor{lightgray} & & LR & 0.06 $\pm$ 0.13 & 0.07 $\pm$ 0.09 & 0.04 $\pm$ 0.09 & 0.04 $\pm$ 0.08 \\
\rowcolor{lightgray} & & RF & 0.09 $\pm$ 0.08 & 0.07 $\pm$ 0.08 & 0.03 $\pm$ 0.09 & 0.04 $\pm$ 0.08 \\
\rowcolor{lightgray} \multirow{-3}{*}{German} &\multirow{-3}{*}{age} & SVC & 0.09 $\pm$ 0.10 & 0.06 $\pm$ 0.09 & 0.04 $\pm$ 0.07 & 0.07 $\pm$ 0.04 \\
NIH &gender &DenseNet & 0.00 $\pm$ 0.00 & 0.01 $\pm$ 0.00** & 0.01 $\pm$ 0.01* & 0.01 $\pm$ 0.01*\\
\rowcolor{lightgray}Fitzpatrick17k &skin type &vgg16 & 0.15 $\pm$ 0.02 & 0.14 $\pm$ 0.02 & 0.11 $\pm$ 0.02** & 0.09 $\pm$ 0.02** \\
\bottomrule
\end{tabular}
\end{table}

\FloatBarrier
\subsection{Label bias}\label{app:bias_f}

\begin{table}[H]\centering
\caption{Balanced accuracy as the percentage of flipped positives in the disadvantaged group varies from $f=0$ (no bias) to $f=1$ (maximum bias).}\label{tab:bal_acc_a_f}
\tiny
\renewcommand{\arraystretch}{1.5} 
\begin{tabular}{lllllll}\toprule
& & & \multicolumn{4}{c}{\textbf{Balanced Accuracy}}\\ \cmidrule(lr){4-7}
\textbf{Dataset} &\textbf{sensitive} &\textbf{model} & \multicolumn{1}{c}{$f=0$ \tiny{(no bias)}} &\multicolumn{1}{c}{$f=0.2$} &\multicolumn{1}{c}{$f=0.8$} &\multicolumn{1}{c}{$f=1$ \tiny{(max bias)}} \\ \midrule
\rowcolor{lightgray} &  & LR & 0.77 $\pm$ 0.00 & 0.76 $\pm$ 0.01* & 0.73 $\pm$ 0.00** & 0.73 $\pm$ 0.00**\\
\rowcolor{lightgray} &  & RF & 0.78 $\pm$ 0.00 & 0.77 $\pm$ 0.01* & 0.74 $\pm$ 0.01** & 0.74 $\pm$ 0.00**\\
\rowcolor{lightgray} & \multirow{-3}{*}{gender} & SVC & 0.77 $\pm$ 0.01 & 0.76 $\pm$ 0.01 & 0.73 $\pm$ 0.01** & 0.73 $\pm$ 0.01**\\
\rowcolor{lightgray}  & & LR & 0.77 $\pm$ 0.00 & 0.76 $\pm$ 0.01* & 0.75 $\pm$ 0.01** & 0.75 $\pm$ 0.01**\\
\rowcolor{lightgray}  & & RF & 0.78 $\pm$ 0.00 & 0.77 $\pm$ 0.00** & 0.75 $\pm$ 0.00** & 0.75 $\pm$ 0.00**\\
\rowcolor{lightgray} \multirow{-6}{*}{Adult} &\multirow{-3}{*}{marital-status} & SVC & 0.77 $\pm$ 0.01 & 0.76 $\pm$ 0.01 & 0.75 $\pm$ 0.01** & 0.75 $\pm$ 0.01**\\
 & & LR & 0.67 $\pm$ 0.02 & 0.68 $\pm$ 0.02 & 0.58 $\pm$ 0.01** & 0.55 $\pm$ 0.01**\\
 & & RF & 0.69 $\pm$ 0.02 & 0.69 $\pm$ 0.02 & 0.59 $\pm$ 0.01** & 0.57 $\pm$ 0.01**\\
\multirow{-3}{*}{Compas} &\multirow{-3}{*}{race} & SVC & 0.65 $\pm$ 0.03 & 0.66 $\pm$ 0.02 & 0.51 $\pm$ 0.01** & 0.50 $\pm$ 0.00**\\
\rowcolor{lightgray} & & LR & 0.84 $\pm$ 0.02 & 0.83 $\pm$ 0.02 & 0.82 $\pm$ 0.02 & 0.81 $\pm$ 0.03\\
\rowcolor{lightgray} & & RF & 0.83 $\pm$ 0.03 & 0.83 $\pm$ 0.02 & 0.82 $\pm$ 0.03 & 0.81 $\pm$ 0.02\\
\rowcolor{lightgray}\multirow{-3}{*}{Crime} &\multirow{-3}{*}{race} & SVC & 0.84 $\pm$ 0.02 & 0.84 $\pm$ 0.02 & 0.82 $\pm$ 0.02 & 0.81 $\pm$ 0.02* \\
 & &LR & 0.73 $\pm$ 0.01 & 0.73 $\pm$ 0.01 & 0.72 $\pm$ 0.01 & 0.72 $\pm$ 0.01 \\
 & &RF & 0.78 $\pm$ 0.00 & 0.78 $\pm$ 0.01 & 0.78 $\pm$ 0.01 & 0.78 $\pm$ 0.01 \\
\multirow{-3}{*}{Folktables} &\multirow{-3}{*}{race} &SVC & 0.72 $\pm$ 0.01 & 0.72 $\pm$ 0.01 & 0.73 $\pm$ 0.01 & 0.73 $\pm$ 0.01 \\
\rowcolor{lightgray}  & & LR & 0.66 $\pm$ 0.06 & 0.68 $\pm$ 0.07 & 0.70 $\pm$ 0.08 & 0.70 $\pm$ 0.08 \\
\rowcolor{lightgray}  & & RF & 0.68 $\pm$ 0.06 & 0.67 $\pm$ 0.05 & 0.72 $\pm$ 0.05 & 0.71 $\pm$ 0.05 \\
\rowcolor{lightgray} \multirow{-3}{*}{German} &\multirow{-3}{*}{age} & SVC & 0.66 $\pm$ 0.07 & 0.66 $\pm$ 0.07 & 0.70 $\pm$ 0.07 & 0.70 $\pm$ 0.06 \\
NIH &gender &DenseNet & 0.64 $\pm$ 0.02& 0.65 $\pm$ 0.01& 0.61 $\pm$ 0.02*& 0.59 $\pm$ 0.02**\\
\rowcolor{lightgray}Fitzpatrick17k &skin type &vgg16 & 0.73 $\pm$ 0.01& 0.71 $\pm$ 0.01**& 0.70 $\pm$ 0.02**& 0.70 $\pm$ 0.02**\\
\bottomrule
\end{tabular}
\end{table}
\begin{table}[H]\centering
\caption{Prediction quality parity (PQP) as the percentage of flipped positives in the disadvantaged group varies from $f=0$ (no bias) to $f=1$ (maximum bias).}\label{tab:pqp_f}
\tiny
\renewcommand{\arraystretch}{1.5} 
\begin{tabular}{lllllll}\toprule
& & & \multicolumn{4}{c}{\textbf{PQP}}\\ \cmidrule(lr){4-7}
\textbf{Dataset} &\textbf{sensitive} &\textbf{model} & \multicolumn{1}{c}{$f=0$ \tiny{(no bias)}} &\multicolumn{1}{c}{$f=0.2$} &\multicolumn{1}{c}{$f=0.8$} &\multicolumn{1}{c}{$f=1$ \tiny{(max bias)}} \\ \midrule
\rowcolor{lightgray} & & LR & \phantom{-}0.00 $\pm$ 0.01\phantom{**} & \phantom{-}0.06 $\pm$ 0.01** & \phantom{-}0.21 $\pm$ 0.01** & \phantom{-}0.23 $\pm$ 0.01** \\
\rowcolor{lightgray} & & RF & \phantom{-}0.01 $\pm$ 0.01 & \phantom{-}0.06 $\pm$ 0.02** & \phantom{-}0.24 $\pm$ 0.01** & \phantom{-}0.25 $\pm$ 0.01**\\
\rowcolor{lightgray} &\multirow{-3}{*}{gender} & SVC & \phantom{-}0.01 $\pm$ 0.01 & \phantom{-}0.05 $\pm$ 0.01** & \phantom{-}0.22 $\pm$ 0.01** & \phantom{-}0.24 $\pm$ 0.01** \\
\rowcolor{lightgray}  & & LR & \phantom{-}0.09 $\pm$ 0.01 & \phantom{-}0.12 $\pm$ 0.02** & \phantom{-}0.21 $\pm$ 0.01** & \phantom{-}0.21 $\pm$ 0.01** \\
\rowcolor{lightgray}  & & RF& \phantom{-}0.08 $\pm$ 0.01 & \phantom{-}0.11 $\pm$ 0.02** & \phantom{-}0.25 $\pm$ 0.01** & \phantom{-}0.25 $\pm$ 0.01** \\
\rowcolor{lightgray} \multirow{-6}{*}{Adult} &\multirow{-3}{*}{marital-status} & SVC & \phantom{-}0.08 $\pm$ 0.02 & \phantom{-}0.11 $\pm$ 0.01** & \phantom{-}0.23 $\pm$ 0.01** & \phantom{-}0.24 $\pm$ 0.01** \\
 & & LR & -0.06 $\pm$ 0.03 & -0.03 $\pm$ 0.05 & \phantom{-}0.06 $\pm$ 0.02** & \phantom{-}0.05 $\pm$ 0.02** \\
 & & RF & -0.01 $\pm$ 0.05 & \phantom{-}0.00 $\pm$ 0.06 & \phantom{-}0.06 $\pm$ 0.03** & \phantom{-}0.05 $\pm$ 0.02** \\
\multirow{-3}{*}{Compas} &\multirow{-3}{*}{race} & SVC & -0.05 $\pm$ 0.03 & -0.04 $\pm$ 0.05 & \phantom{-}0.01 $\pm$ 0.01** & \phantom{-}0.00 $\pm$ 0.00** \\
\rowcolor{lightgray} & & LR & -0.02 $\pm$ 0.08 & \phantom{-}0.01 $\pm$ 0.07 & \phantom{-}0.11 $\pm$ 0.10* & \phantom{-}0.13 $\pm$ 0.08** \\
\rowcolor{lightgray} & & RF & -0.03 $\pm$ 0.05 & \phantom{-}0.02 $\pm$ 0.06 & \phantom{-}0.16 $\pm$ 0.07** & \phantom{-}0.17 $\pm$ 0.06** \\
\rowcolor{lightgray}\multirow{-3}{*}{Crime} &\multirow{-3}{*}{race} & SVC & -0.03 $\pm$ 0.08 & \phantom{-}0.01 $\pm$ 0.08 & \phantom{-}0.10 $\pm$ 0.07** & \phantom{-}0.14 $\pm$ 0.07** \\
 & &LR & \phantom{-}0.01 $\pm$ 0.04 & \phantom{-}0.01 $\pm$ 0.04 & \phantom{-}0.02 $\pm$ 0.04 & \phantom{-}0.02 $\pm$ 0.04\\
 & &RF & \phantom{-}0.01 $\pm$ 0.03 & \phantom{-}0.00 $\pm$ 0.02 & \phantom{-}0.00 $\pm$ 0.03 & \phantom{-}0.01 $\pm$ 0.03\\
\multirow{-3}{*}{Folktables} &\multirow{-3}{*}{race} &SVC & \phantom{-}0.02 $\pm$ 0.03 & \phantom{-}0.02 $\pm$ 0.04 & \phantom{-}0.02 $\pm$ 0.04 & \phantom{-}0.02 $\pm$ 0.04\\
\rowcolor{lightgray} & & LR & \phantom{-}0.07 $\pm$ 0.13 & \phantom{-}0.06 $\pm$ 0.15 & \phantom{-}0.05 $\pm$ 0.16 & \phantom{-}0.06 $\pm$ 0.14 \\
\rowcolor{lightgray} & & RF & -0.01 $\pm$ 0.07 & \phantom{-}0.05 $\pm$ 0.09 & \phantom{-}0.07 $\pm$ 0.09 & \phantom{-}0.09 $\pm$ 0.14 \\
\rowcolor{lightgray} \multirow{-3}{*}{German} &\multirow{-3}{*}{age} & SVC & \phantom{-}0.03 $\pm$ 0.10 & \phantom{-}0.02 $\pm$ 0.10 & \phantom{-}0.02 $\pm$ 0.11 & \phantom{-}0.03 $\pm$ 0.15 \\
NIH &gender &DenseNet & -0.01 $\pm$ 0.01 & \phantom{-}0.01 $\pm$ 0.01** & \phantom{-}0.10 $\pm$ 0.01**  & \phantom{-}0.14 $\pm$ 0.01**\\
\rowcolor{lightgray}Fitzpatrick17k &skin type &vgg16 & -0.01 $\pm$ 0.01 & \phantom{-}0.01 $\pm$ 0.01** & \phantom{-}0.20 $\pm$ 0.02** & \phantom{-}0.05 $\pm$ 0.01**\\
\bottomrule
\end{tabular}
\end{table}

\begin{table}[H]\centering
\caption{True positive rate (TPR) of the advantaged group as the percentage of flipped positives in the disadvantaged group varies from $f=0$ (no bias) to $f=1$ (maximum bias).}\label{tab:tpr_a_f}
\tiny
\renewcommand{\arraystretch}{1.5} 
\begin{tabular}{lllllll}\toprule
& & & \multicolumn{4}{c}{\textbf{TPR ($s=a$)}}\\ \cmidrule(lr){4-7}
\textbf{Dataset} &\textbf{sensitive} &\textbf{model} &\multicolumn{1}{c}{$f=0$ \tiny{(no bias)}} & \multicolumn{1}{c}{$f=0.2$} & \multicolumn{1}{c}{$f=0.8$} & \multicolumn{1}{c}{$f=1$ \tiny{(max bias)}}\\ \midrule
\rowcolor{lightgray} & & LR & 0.61 $\pm$ 0.01\phantom{**} & 0.61 $\pm$ 0.01\phantom{**} & 0.60 $\pm$ 0.01\phantom{**} & 0.60 $\pm$ 0.01\phantom{**}\\
\rowcolor{lightgray} & & RF & 0.64 $\pm$ 0.01 & 0.63 $\pm$ 0.01 & 0.62 $\pm$ 0.01** & 0.62 $\pm$ 0.01**\\
\rowcolor{lightgray} &\multirow{-3}{*}{gender} & SVC & 0.61 $\pm$ 0.01 & 0.60 $\pm$ 0.01 & 0.60 $\pm$ 0.01 & 0.59 $\pm$ 0.01**\\
\rowcolor{lightgray} & & LR & 0.65 $\pm$ 0.01 & 0.65 $\pm$ 0.01 & 0.65 $\pm$ 0.01 & 0.65 $\pm$ 0.01 \\
\rowcolor{lightgray} & & RF & 0.66 $\pm$ 0.01 & 0.66 $\pm$ 0.01 & 0.65 $\pm$ 0.01 & 0.66 $\pm$ 0.01 \\
\rowcolor{lightgray} \multirow{-6}{*}{Adult} &\multirow{-3}{*}{marital-status} & SVC & 0.64 $\pm$ 0.01 & 0.64 $\pm$ 0.01 & 0.65 $\pm$ 0.01 & 0.65 $\pm$ 0.01 \\
 & & LR & 0.85 $\pm$ 0.02 & 0.76 $\pm$ 0.03** & 0.33 $\pm$ 0.04** & 0.22 $\pm$ 0.04**\\
 & & RF & 0.81 $\pm$ 0.03 & 0.72 $\pm$ 0.03** & 0.34 $\pm$ 0.05** & 0.25 $\pm$ 0.05**\\
\multirow{-3}{*}{Compas} &\multirow{-3}{*}{race} & SVC & 0.85 $\pm$ 0.04 & 0.79 $\pm$ 0.03** & 0.03 $\pm$ 0.06** & 0.00 $\pm$ 0.00**\\
\rowcolor{lightgray} & & LR & 0.90 $\pm$ 0.04 & 0.89 $\pm$ 0.04 & 0.87 $\pm$ 0.04 & 0.85 $\pm$ 0.05 \\
\rowcolor{lightgray} & & RF & 0.90 $\pm$ 0.03 & 0.89 $\pm$ 0.03 & 0.86 $\pm$ 0.03** & 0.85 $\pm$ 0.02** \\
\rowcolor{lightgray}\multirow{-3}{*}{Crime} &\multirow{-3}{*}{race} & SVC & 0.91 $\pm$ 0.04 & 0.90 $\pm$ 0.04 & 0.88 $\pm$ 0.04 & 0.87 $\pm$ 0.04 \\
 & & LR & 0.83 $\pm$ 0.01 & 0.82 $\pm$ 0.01 & 0.77 $\pm$ 0.01** & 0.75 $\pm$ 0.01** \\
 & & RF & 0.85 $\pm$ 0.01 & 0.85 $\pm$ 0.01 & 0.83 $\pm$ 0.02** & 0.83 $\pm$ 0.02** \\
\multirow{-3}{*}{Folktables} &\multirow{-3}{*}{race} &SVC & 0.85 $\pm$ 0.01 & 0.83 $\pm$ 0.01** & 0.78 $\pm$ 0.01** & 0.76 $\pm$ 0.01** \\
\rowcolor{lightgray} & & LR & 0.89 $\pm$ 0.05 & 0.88 $\pm$ 0.05 & 0.82 $\pm$ 0.05** & 0.79 $\pm$ 0.06** \\
\rowcolor{lightgray} & & RF & 0.93 $\pm$ 0.02 & 0.91 $\pm$ 0.04 & 0.86 $\pm$ 0.05* & 0.82 $\pm$ 0.05* \\
\rowcolor{lightgray} \multirow{-3}{*}{German} &\multirow{-3}{*}{age} & SVC & 0.93 $\pm$ 0.02 & 0.89 $\pm$ 0.06 & 0.83 $\pm$ 0.05** & 0.82 $\pm$ 0.06** \\
NIH &gender &DenseNet & 0.46 $\pm$ 0.03 & 0.49 $\pm$ 0.03 & 0.59 $\pm$ 0.06** & 0.52 $\pm$ 0.05* \\
\rowcolor{lightgray}Fitzpatrick17k &skin type &vgg16 & 0.70 $\pm$ 0.02 & 0.70 $\pm$ 0.02 & 0.69 $\pm$ 0.02 & 0.70 $\pm$ 0.01\\
\bottomrule
\end{tabular}
\end{table}

\begin{table}[H]\centering
\caption{True positive rate (TPR) of the disadvantaged group as the percentage of flipped positives in the disadvantaged group varies from $f=0$ (no bias) to $f=1$ (maximum bias).}\label{tab:tpr_d_f}
\tiny
\renewcommand{\arraystretch}{1.5} 
\begin{tabular}{lllllll}\toprule
& & & \multicolumn{4}{c}{\textbf{TPR ($s = d$)}}\\ \cmidrule(lr){4-7}
\textbf{Dataset} &\textbf{sensitive} &\textbf{model} &\multicolumn{1}{c}{$f=0$ \tiny{(no bias)}} & \multicolumn{1}{c}{$f=0.2$} & \multicolumn{1}{c}{$f=0.8$} & \multicolumn{1}{c}{$f=1$ \tiny{(max bias)}}\\ \midrule
\rowcolor{lightgray} & & LR & 0.53 $\pm$ 0.02\phantom{**} & 0.40 $\pm$ 0.03** & 0.07 $\pm$ 0.02** & 0.05 $\pm$ 0.02**\\
\rowcolor{lightgray} & & RF & 0.54 $\pm$ 0.02 & 0.43 $\pm$ 0.04** & 0.06 $\pm$ 0.01** & 0.04 $\pm$ 0.01**\\
\rowcolor{lightgray} &\multirow{-3}{*}{gender} & SVC & 0.53 $\pm$ 0.02 & 0.42 $\pm$ 0.03** & 0.06 $\pm$ 0.02** & 0.03 $\pm$ 0.01**\\
\rowcolor{lightgray} & & LR & 0.29 $\pm$ 0.03 & 0.22 $\pm$ 0.04** & 0.03 $\pm$ 0.01** & 0.02 $\pm$ 0.01** \\
\rowcolor{lightgray} & & RF & 0.35 $\pm$ 0.03 & 0.29 $\pm$ 0.03** & 0.00 $\pm$ 0.00** & 0.00 $\pm$ 0.00** \\
\rowcolor{lightgray} \multirow{-6}{*}{Adult} &\multirow{-3}{*}{marital-status} & SVC & 0.30 $\pm$ 0.03 & 0.26 $\pm$ 0.03* & 0.01 $\pm$ 0.01** & 0.00 $\pm$ 0.00** \\
 & & LR & 0.68 $\pm$ 0.03 & 0.55 $\pm$ 0.05** & 0.11 $\pm$ 0.02** & 0.07 $\pm$ 0.02**\\
 & & RF & 0.66 $\pm$ 0.03 & 0.57 $\pm$ 0.04** & 0.16 $\pm$ 0.02** & 0.09 $\pm$ 0.01**\\
\multirow{-3}{*}{Compas} &\multirow{-3}{*}{race} & SVC & 0.68 $\pm$ 0.06 & 0.58 $\pm$ 0.05** & 0.01 $\pm$ 0.02** & 0.00 $\pm$ 0.00**\\
\rowcolor{lightgray} & & LR & 0.57 $\pm$ 0.10 & 0.50 $\pm$ 0.11 & 0.25 $\pm$ 0.15** & 0.18 $\pm$ 0.12** \\
\rowcolor{lightgray} & & RF & 0.57 $\pm$ 0.07 & 0.47 $\pm$ 0.10 & 0.19 $\pm$ 0.12** & 0.18 $\pm$ 0.10** \\
\rowcolor{lightgray} \multirow{-3}{*}{Crime} &\multirow{-3}{*}{race} & SVC & 0.59 $\pm$ 0.08 & 0.50 $\pm$ 0.11 & 0.26 $\pm$ 0.11** & 0.18 $\pm$ 0.10** \\
 & &LR & 0.79 $\pm$ 0.04 & 0.77 $\pm$ 0.04 & 0.69 $\pm$ 0.04** & 0.66 $\pm$ 0.04** \\
 & &RF & 0.83 $\pm$ 0.03 & 0.82 $\pm$ 0.03 & 0.77 $\pm$ 0.05* & 0.77 $\pm$ 0.05* \\
\multirow{-3}{*}{Folktables} &\multirow{-3}{*}{race} &SVC & 0.80 $\pm$ 0.04 & 0.78 $\pm$ 0.04 & 0.70 $\pm$ 0.04** & 0.67 $\pm$ 0.04** \\
\rowcolor{lightgray} & & LR & 0.82 $\pm$ 0.13 & 0.78 $\pm$ 0.15 & 0.60 $\pm$ 0.17* & 0.54 $\pm$ 0.13** \\
\rowcolor{lightgray} & & RF & 0.90 $\pm$ 0.07 & 0.80 $\pm$ 0.09** & 0.60 $\pm$ 0.10** & 0.50 $\pm$ 0.16** \\
\rowcolor{lightgray} \multirow{-3}{*}{German} &\multirow{-3}{*}{age} & SVC & 0.87 $\pm$ 0.09 & 0.83 $\pm$ 0.14 & 0.61 $\pm$ 0.11** & 0.55 $\pm$ 0.11** \\
NIH &gender &DenseNet & 0.44 $\pm$ 0.05 & 0.40 $\pm$ 0.04 & 0.18 $\pm$ 0.07** & 0.03 $\pm$ 0.01** \\
\rowcolor{lightgray}Fitzpatrick17k &skin type &vgg16 & 0.61 $\pm$ 0.05 & 0.53 $\pm$ 0.06* & 0.48 $\pm$ 0.08** & 0.42 $\pm$ 0.05** \\
\bottomrule
\end{tabular}
\end{table}

\begin{table}[H]\centering
\caption{Equal opportunity (EO) as the percentage of flipped positives in the disadvantaged group varies from $f=0$ (no bias) to $f=1$ (maximum bias).}\label{tab:eo_f}
\tiny
\renewcommand{\arraystretch}{1.5} 
\begin{tabular}{lllllll}\toprule
& & & \multicolumn{4}{c}{\textbf{EO}}\\ \cmidrule(lr){4-7}
\textbf{Dataset} &\textbf{sensitive} &\textbf{model} &\multicolumn{1}{c}{$f=0$ \tiny{(no bias)}} & \multicolumn{1}{c}{$f=0.2$} & \multicolumn{1}{c}{$f=0.8$} & \multicolumn{1}{c}{$f=1$ \tiny{(max bias)}}\\ \midrule
\rowcolor{lightgray} & & LR & 0.08 $\pm$ 0.02\phantom{**} & 0.21 $\pm$ 0.02** & 0.52 $\pm$ 0.03** & 0.55 $\pm$ 0.02**\\
\rowcolor{lightgray} & & RF & 0.10 $\pm$ 0.02 & 0.20 $\pm$ 0.04** & 0.56 $\pm$ 0.01** & 0.57 $\pm$ 0.02**\\
\rowcolor{lightgray} &\multirow{-3}{*}{gender} & SVC & 0.08 $\pm$ 0.02 & 0.18 $\pm$ 0.02** & 0.54 $\pm$ 0.02** & 0.56 $\pm$ 0.01**\\
\rowcolor{lightgray} & & LR & 0.36 $\pm$ 0.03 & 0.43 $\pm$ 0.04** & 0.62 $\pm$ 0.02** & 0.63 $\pm$ 0.02**\\
\rowcolor{lightgray} & & RF & 0.31 $\pm$ 0.03 & 0.37 $\pm$ 0.03** & 0.65 $\pm$ 0.01** & 0.66 $\pm$ 0.01**\\
\rowcolor{lightgray} \multirow{-6}{*}{Adult} &\multirow{-3}{*}{marital-status} & SVC & 0.34 $\pm$ 0.03 & 0.39 $\pm$ 0.02** & 0.65 $\pm$ 0.01** & 0.65 $\pm$ 0.01**\\
 & & LR & 0.17 $\pm$ 0.03 & 0.21 $\pm$ 0.05 & 0.21 $\pm$ 0.05 & 0.15 $\pm$ 0.05 \\
 & & RF & 0.15 $\pm$ 0.05 & 0.15 $\pm$ 0.06 & 0.18 $\pm$ 0.05 & 0.15 $\pm$ 0.05 \\
\multirow{-3}{*}{Compas} &\multirow{-3}{*}{race} & SVC & 0.18 $\pm$ 0.04 & 0.21 $\pm$ 0.04 & 0.02 $\pm$ 0.04** & 0.00 $\pm$ 0.00** \\
\rowcolor{lightgray} & & LR & 0.33 $\pm$ 0.11 & 0.39 $\pm$ 0.11 & 0.62 $\pm$ 0.17** & 0.67 $\pm$ 0.15**\\
\rowcolor{lightgray} & & RF & 0.33 $\pm$ 0.06 & 0.42 $\pm$ 0.12 & 0.66 $\pm$ 0.13** & 0.67 $\pm$ 0.10**\\
\rowcolor{lightgray} \multirow{-3}{*}{Crime} &\multirow{-3}{*}{race} & SVC & 0.32 $\pm$ 0.08 & 0.40 $\pm$ 0.12 & 0.62 $\pm$ 0.12** & 0.69 $\pm$ 0.12**\\
 & &LR & 0.05 $\pm$ 0.04 & 0.05 $\pm$ 0.04 & 0.08 $\pm$ 0.04 & 0.09 $\pm$ 0.04\\
 & &RF & 0.02 $\pm$ 0.04 & 0.02 $\pm$ 0.03 & 0.05 $\pm$ 0.04 & 0.06 $\pm$ 0.03\\
\multirow{-3}{*}{Folktables} &\multirow{-3}{*}{race} &SVC & 0.04 $\pm$ 0.03 & 0.06 $\pm$ 0.04 & 0.08 $\pm$ 0.03* & 0.09 $\pm$ 0.04*\\
\rowcolor{lightgray} & & LR& 0.07 $\pm$ 0.13 & 0.10 $\pm$ 0.14 & 0.22 $\pm$ 0.16 & 0.25 $\pm$ 0.10**\\
\rowcolor{lightgray} & & RF & 0.03 $\pm$ 0.06 & 0.11 $\pm$ 0.08 & 0.26 $\pm$ 0.09** & 0.32 $\pm$ 0.15** \\
\rowcolor{lightgray} \multirow{-3}{*}{German} &\multirow{-3}{*}{age} & SVC & 0.06 $\pm$ 0.08 & 0.06 $\pm$ 0.11 & 0.22 $\pm$ 0.10** & 0.26 $\pm$ 0.13**\\
NIH &gender &DenseNet & 0.01 $\pm$ 0.02 & 0.09 $\pm$ 0.02** & 0.40 $\pm$ 0.03** & 0.48 $\pm$ 0.01**\\
\rowcolor{lightgray}Fitzpatrick17k &skin type &vgg16 & 0.09 $\pm$ 0.05 & 0.16 $\pm$ 0.06* & 0.21 $\pm$ 0.08** & 0.28 $\pm$ 0.02**\\
\bottomrule
\end{tabular}
\end{table}

\begin{table}[H]\centering
\caption{Demographic parity (DP) as the percentage of flipped positives in the disadvantaged group varies from $f=0$ (no bias) to $f=1$ (maximum bias).}\label{tab:dp_f}
\tiny
\renewcommand{\arraystretch}{1.5} 
\begin{tabular}{lllllll}\toprule
& & & \multicolumn{4}{c}{\textbf{DP}}\\ \cmidrule(lr){4-7}
\textbf{Dataset} &\textbf{sensitive} &\textbf{model} &\multicolumn{1}{c}{$f=0$ \tiny{(no bias)}} & \multicolumn{1}{c}{$f=0.2$} & \multicolumn{1}{c}{$f=0.8$} & \multicolumn{1}{c}{$f=1$ \tiny{(max bias)}} \\ \midrule
\rowcolor{lightgray} & & LR & 0.18 $\pm$ 0.01\phantom{**} & 0.20 $\pm$ 0.01** & 0.25 $\pm$ 0.01** & 0.25 $\pm$ 0.01** \\
\rowcolor{lightgray} & & RF & 0.18 $\pm$ 0.01 & 0.20 $\pm$ 0.01** & 0.24 $\pm$ 0.01** & 0.24 $\pm$ 0.01** \\
\rowcolor{lightgray} &\multirow{-3}{*}{gender} & SVC & 0.17 $\pm$ 0.00 & 0.19 $\pm$ 0.01** & 0.24 $\pm$ 0.01** & 0.24 $\pm$ 0.01** \\
\rowcolor{lightgray}  & & LR & 0.37 $\pm$ 0.01 & 0.38 $\pm$ 0.01 & 0.40 $\pm$ 0.01** & 0.40 $\pm$ 0.01** \\
\rowcolor{lightgray}  & & RF & 0.36 $\pm$ 0.02 & 0.36 $\pm$ 0.02 & 0.38 $\pm$ 0.01* & 0.38 $\pm$ 0.02 \\
\rowcolor{lightgray} \multirow{-6}{*}{Adult} &\multirow{-3}{*}{marital-status} & SVC & 0.36 $\pm$ 0.01 & 0.37 $\pm$ 0.01 & 0.39 $\pm$ 0.01** & 0.39 $\pm$ 0.01** \\
 & & LR & 0.26 $\pm$ 0.03 & 0.28 $\pm$ 0.02 & 0.18 $\pm$ 0.04** & 0.12 $\pm$ 0.03** \\
 & & RF & 0.21 $\pm$ 0.05 & 0.20 $\pm$ 0.05 & 0.16 $\pm$ 0.04 & 0.12 $\pm$ 0.04** \\
\multirow{-3}{*}{Compas} &\multirow{-3}{*}{race} & SVC & 0.26 $\pm$ 0.02 & 0.28 $\pm$ 0.02 & 0.02 $\pm$ 0.03** & 0.00 $\pm$ 0.00** \\
\rowcolor{lightgray}  & & LR & 0.62 $\pm$ 0.06 & 0.63 $\pm$ 0.06 & 0.69 $\pm$ 0.07 & 0.70 $\pm$ 0.08 \\
\rowcolor{lightgray}  & & RF & 0.62 $\pm$ 0.07 & 0.64 $\pm$ 0.07 & 0.67 $\pm$ 0.06 & 0.67 $\pm$ 0.06 \\
\rowcolor{lightgray} \multirow{-3}{*}{Crime} &\multirow{-3}{*}{race} & SVC & 0.62 $\pm$ 0.07 & 0.64 $\pm$ 0.07 & 0.70 $\pm$ 0.06* & 0.70 $\pm$ 0.07 \\
 & &LR & 0.06 $\pm$ 0.03 & 0.07 $\pm$ 0.03 & 0.09 $\pm$ 0.03 & 0.09 $\pm$ 0.03\\
 & &RF & 0.05 $\pm$ 0.03 & 0.06 $\pm$ 0.03 & 0.09 $\pm$ 0.03* & 0.09 $\pm$ 0.04\\
\multirow{-3}{*}{Folktables} &\multirow{-3}{*}{race} &SVC & 0.06 $\pm$ 0.03 & 0.06 $\pm$ 0.03 & 0.09 $\pm$ 0.03 & 0.09 $\pm$ 0.03\\
\rowcolor{lightgray}  & & LR & 0.06 $\pm$ 0.13 & 0.11 $\pm$ 0.13 & 0.24 $\pm$ 0.16* & 0.27 $\pm$ 0.15** \\
\rowcolor{lightgray}  & & RF & 0.09 $\pm$ 0.08 & 0.12 $\pm$ 0.08 & 0.28 $\pm$ 0.12** & 0.31 $\pm$ 0.11** \\
\rowcolor{lightgray} \multirow{-3}{*}{German} &\multirow{-3}{*}{age} & SVC & 0.09 $\pm$ 0.10 & 0.09 $\pm$ 0.11 & 0.26 $\pm$ 0.14* & 0.30 $\pm$ 0.11** \\
NIH &gender &DenseNet & 0.00 $\pm$ 0.00 & 0.02 $\pm$ 0.00** & 0.10 $\pm$ 0.02** & 0.10 $\pm$ 0.01**\\
\rowcolor{lightgray}Fitzpatrick17k &skin type &vgg16 & 0.15 $\pm$ 0.02 & 0.17 $\pm$ 0.02 & 0.22 $\pm$ 0.02** & 0.23 $\pm$ 0.02** \\
\bottomrule
\end{tabular}
\end{table}

\FloatBarrier

\FloatBarrier
\section{Additional Results on Bias Detection} \label{app:exp_bias_detection}

In this section, we report bias detection results regarding all datasets except Folktables and NIH, which are discussed in Section \ref{sec:bias_detection}.

\begin{figure}
    \centering
    \includegraphics[width=\linewidth]{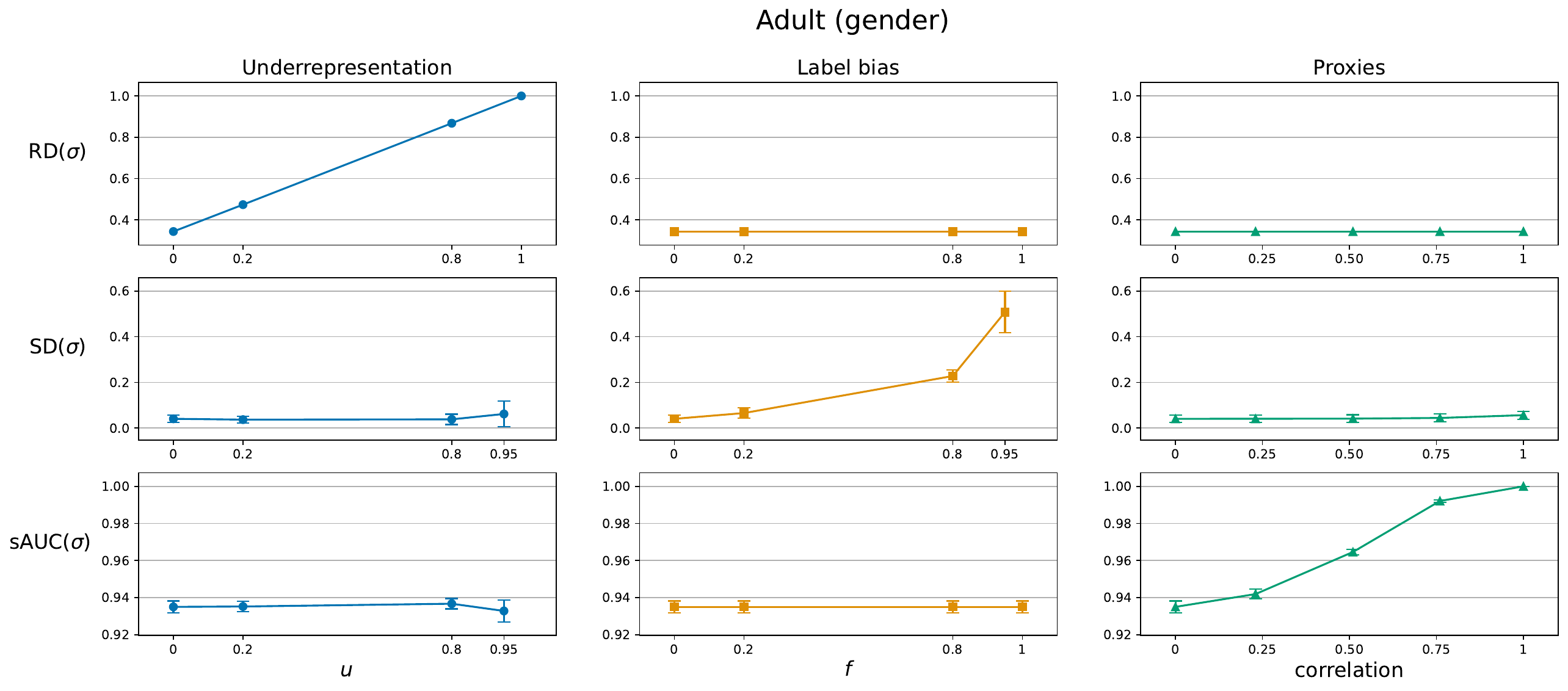}
    \caption{Results of the bias detection methods on Adult (gender). The columns represent the three different scenarios while the rows represent the three bias detection methods.}
    \label{fig:adult_gender_bias_detection}
\end{figure}

\begin{figure}
    \centering
    \includegraphics[width=\linewidth]{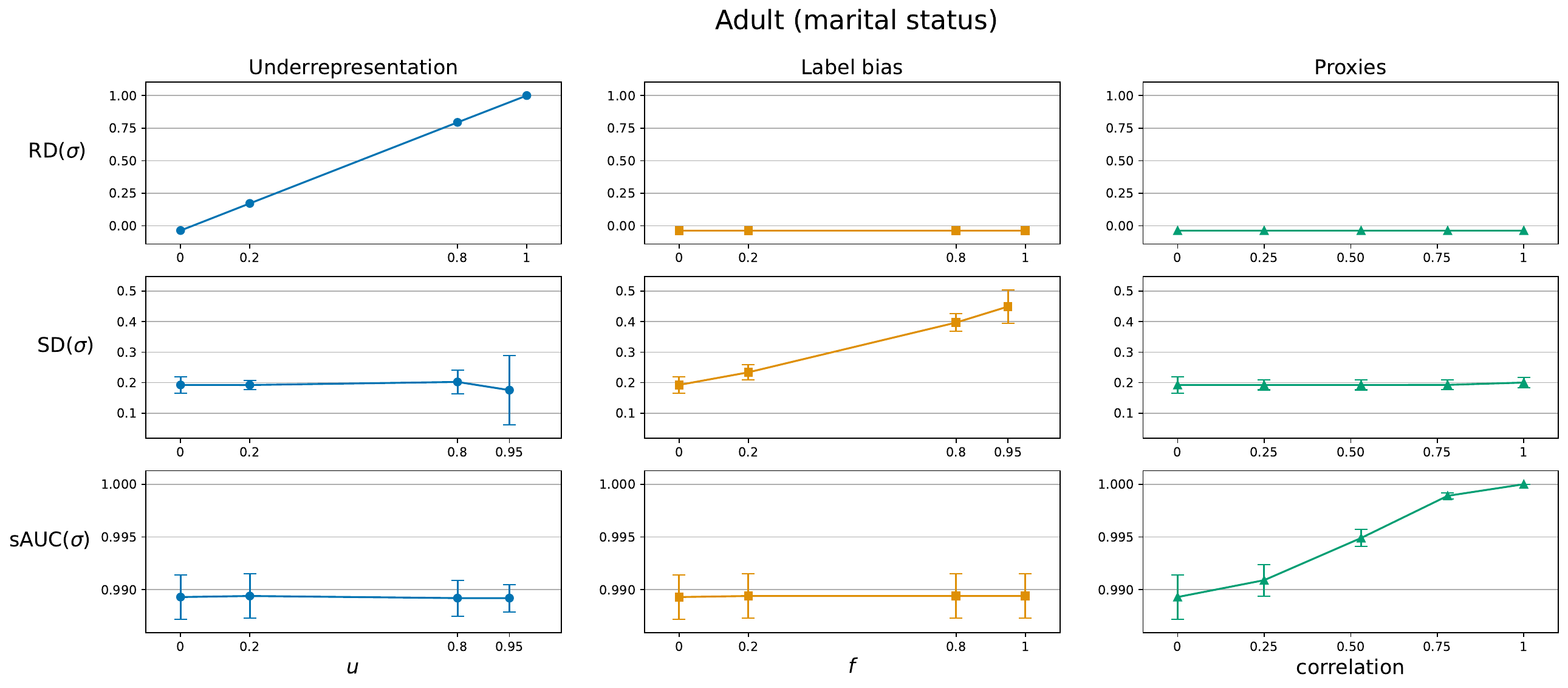}
    \caption{Results of the bias detection methods on Adult (marital status). The columns represent the three different scenarios while the rows represent the three bias detection methods.}
    \label{fig:adult_ms_bias_detection}
\end{figure}

\begin{figure}
    \centering
    \includegraphics[width=\linewidth]{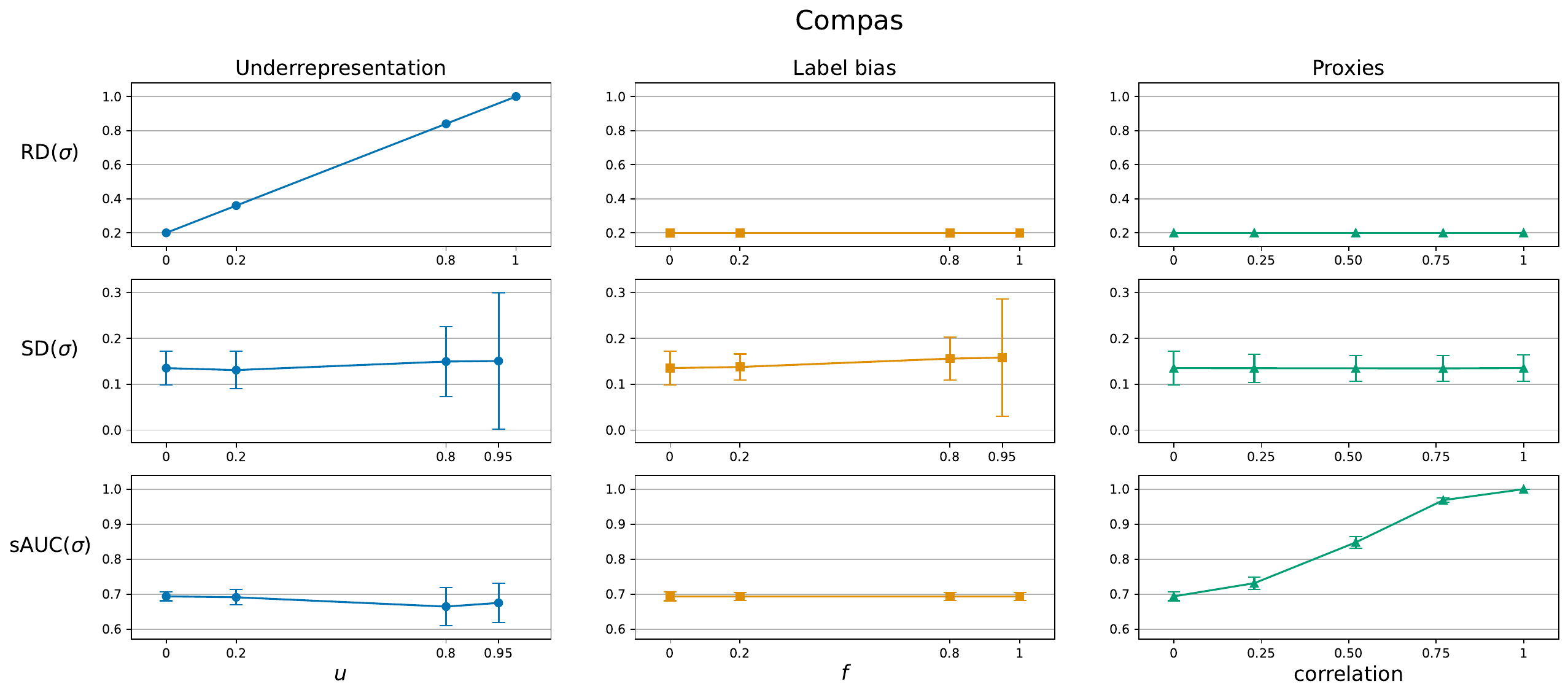}
    \caption{Results of the bias detection methods on Compas. The columns represent the three different scenarios while the rows represent the three bias detection methods.}
    \label{fig:compas_bias_detection}
\end{figure}

\begin{figure}
    \centering
    \includegraphics[width=\linewidth]{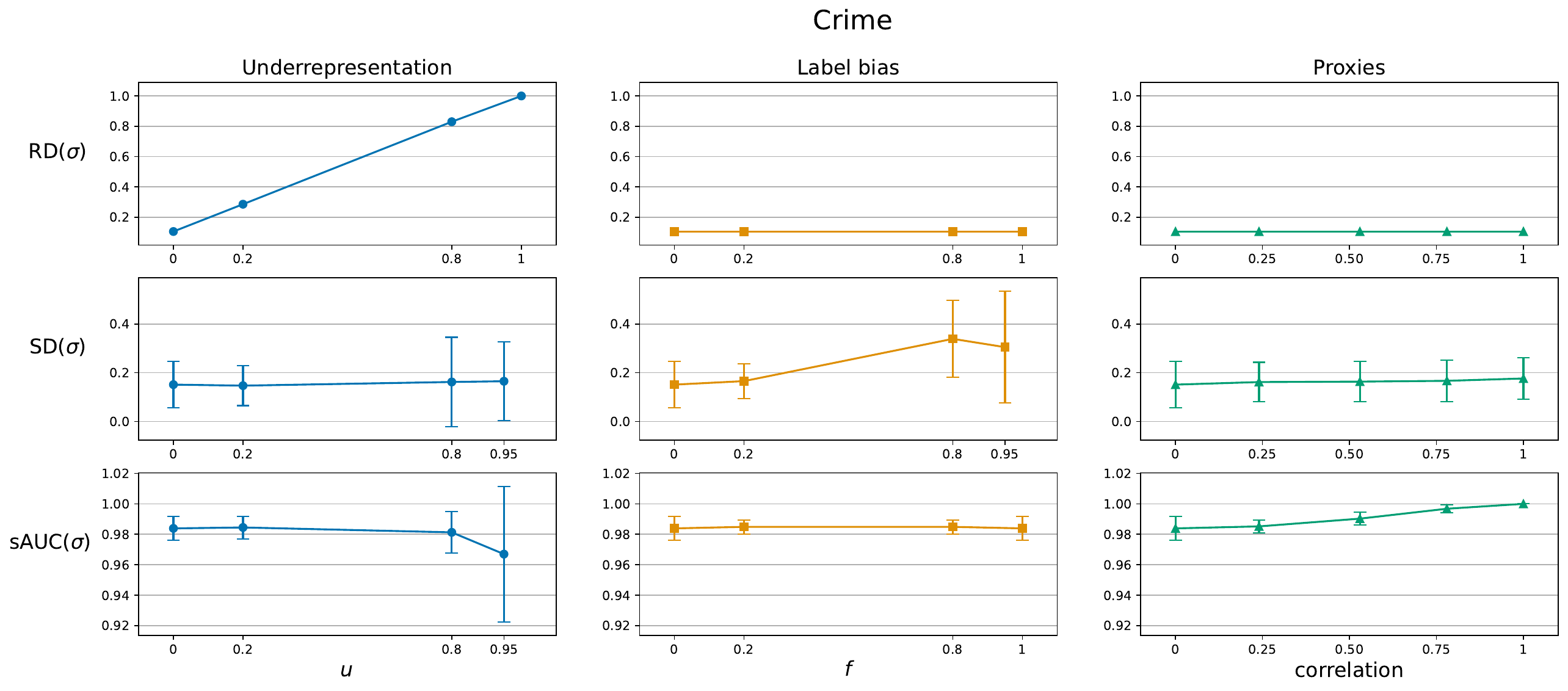}
    \caption{Results of the bias detection methods on Crime. The columns represent the three different scenarios while the rows represent the three bias detection methods.}
    \label{fig:crime_bias_detection}
\end{figure}

\begin{figure}
    \centering
    \includegraphics[width=\linewidth]{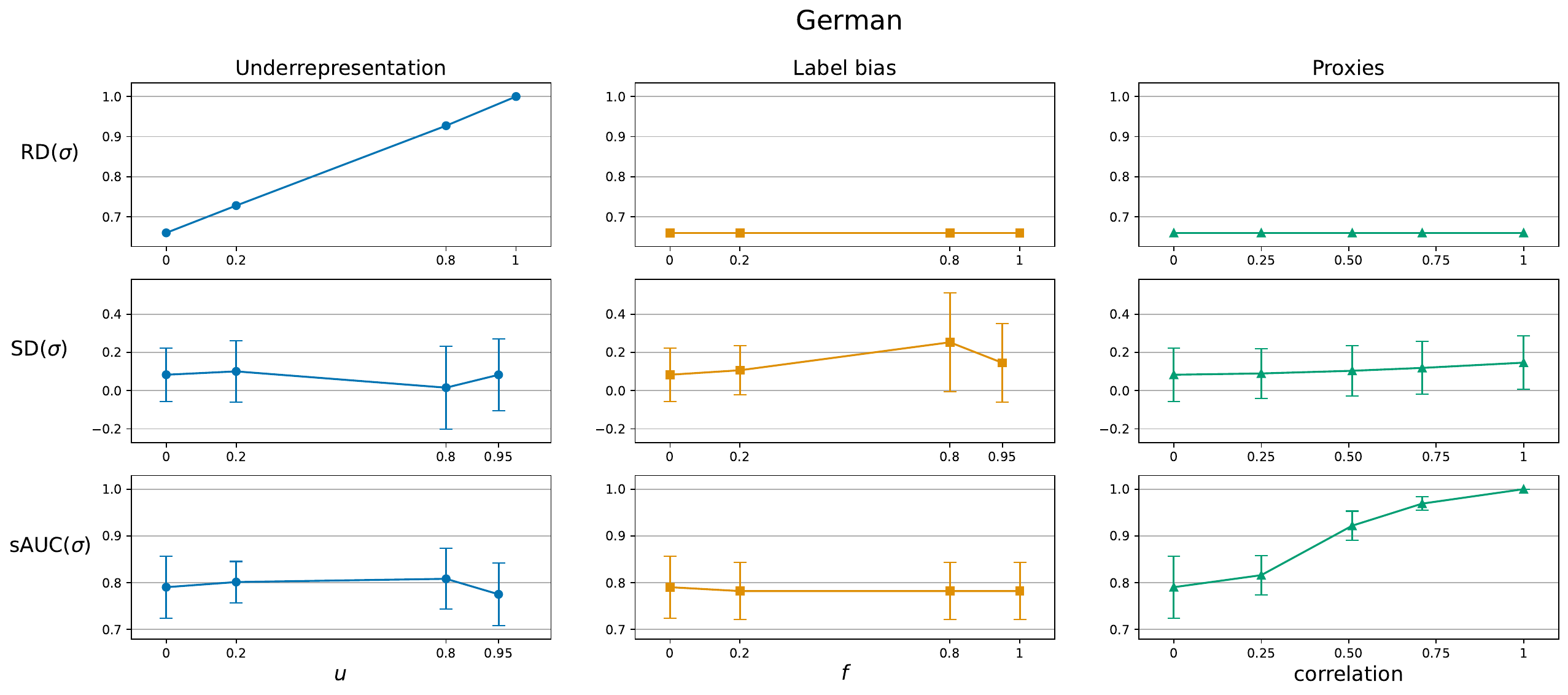}
    \caption{Results of the bias detection methods on German. The columns represent the three different scenarios while the rows represent the three bias detection methods.}
    \label{fig:german_bias_detection}
\end{figure}

\begin{figure}
    \centering
    \includegraphics[width=\linewidth]{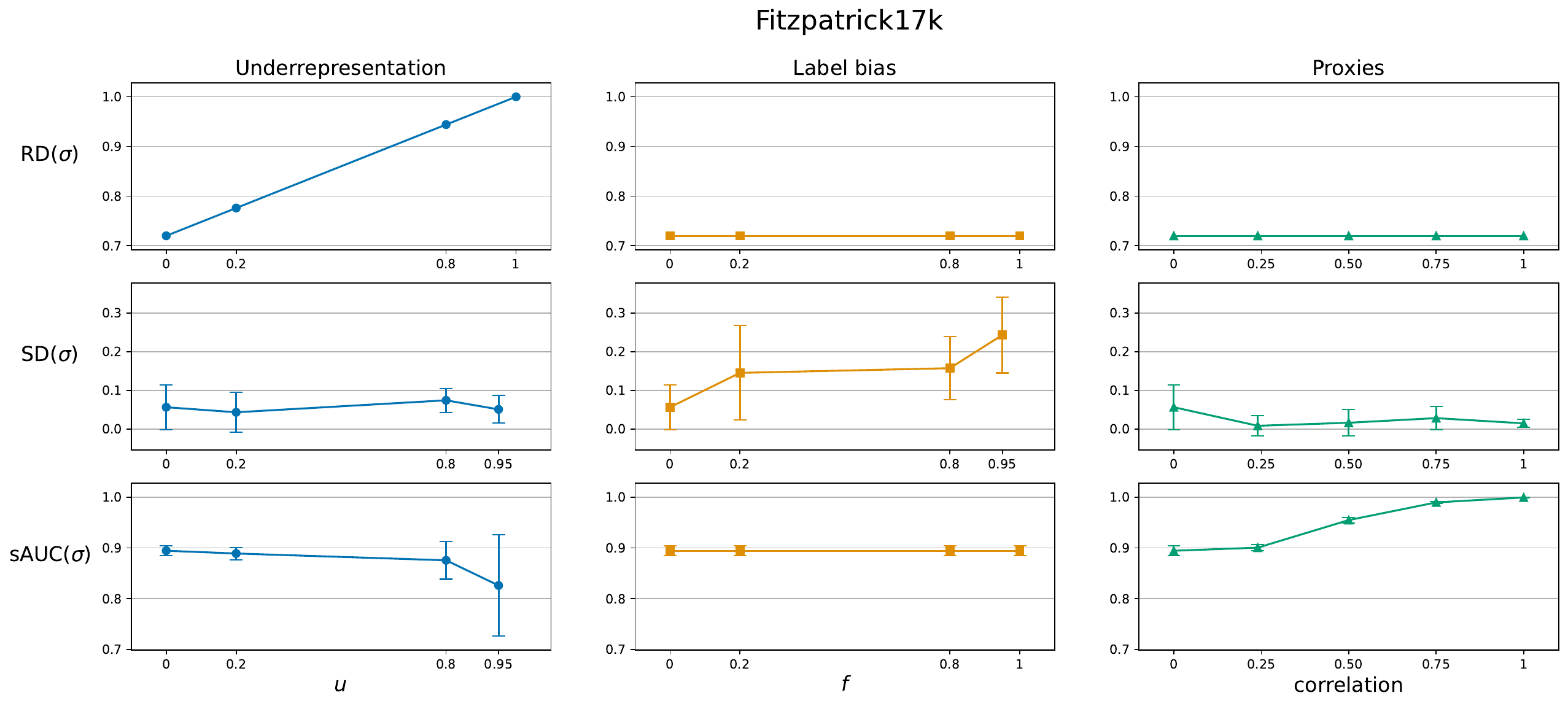}
    \caption{Results of the bias detection methods on Fitzpatrick17k. The columns represent the three different scenarios while the rows represent the three bias detection methods.}
    \label{fig:fitzpatrick_bias_detection}
\end{figure}

\clearpage

\bibliographystyle{unsrtnat}  
\bibliography{biblio}

\end{document}